\pdfoutput=1
\documentclass[11pt]{article}
\usepackage[utf8]{inputenc}
\usepackage{newunicodechar}
\newunicodechar{∞}{$\infty$}

\usepackage[preprint]{acl}
\usepackage{flushend}
\usepackage{listings}
\usepackage{xcolor}
\usepackage{verbatim}
\usepackage{float}
\definecolor{lightgray}{rgb}{0.95,0.95,0.95}

\definecolor{codeblue}{rgb}{0.1,0.1,0.6}
\definecolor{keygray}{rgb}{0.5,0.0,0.0}

\lstdefinestyle{jsonstyle}{
    backgroundcolor=\color{lightgray},cmark
    basicstyle=\footnotesize\ttfamily,
    keywordstyle=\color{keygray}\bfseries,
    stringstyle=\color{codeblue},
    numbers=none,
    showstringspaces=false,
    breaklines=true,
    frame=single
}

\usepackage{times}
\usepackage{makecell}
\usepackage{latexsym}
\usepackage{tcolorbox}
\usepackage{tabularx}
\usepackage{amsmath}
\usepackage[T1]{fontenc}
\usepackage{pdfpages}
\usepackage{multirow}
\usepackage{booktabs}
\usepackage{amssymb}
\usepackage{pifont}
\usepackage{hyperref}
\usepackage{fancyvrb}
\usepackage{fontawesome}
\usepackage{graphicx}
\usepackage{xcolor}
\usepackage{rotating}
\usepackage{array}
\usepackage{longtable}
\usepackage{enumitem}
\usepackage{algorithm}
\usepackage{algpseudocode}
\usepackage{url}
\usepackage{microtype}
\usepackage{inconsolata}
\usepackage[utf8]{inputenc}
\usepackage{semantic}
\usepackage{color}
\usepackage{amssymb}

\newcommand{\cmark}{\textcolor{green!60!black}{\ding{51}}}
\newcommand{\xmark}{\textcolor{red}{\ding{55}}}

\NewDocumentCommand\emojismile{}{
    \includegraphics[scale=0.05]{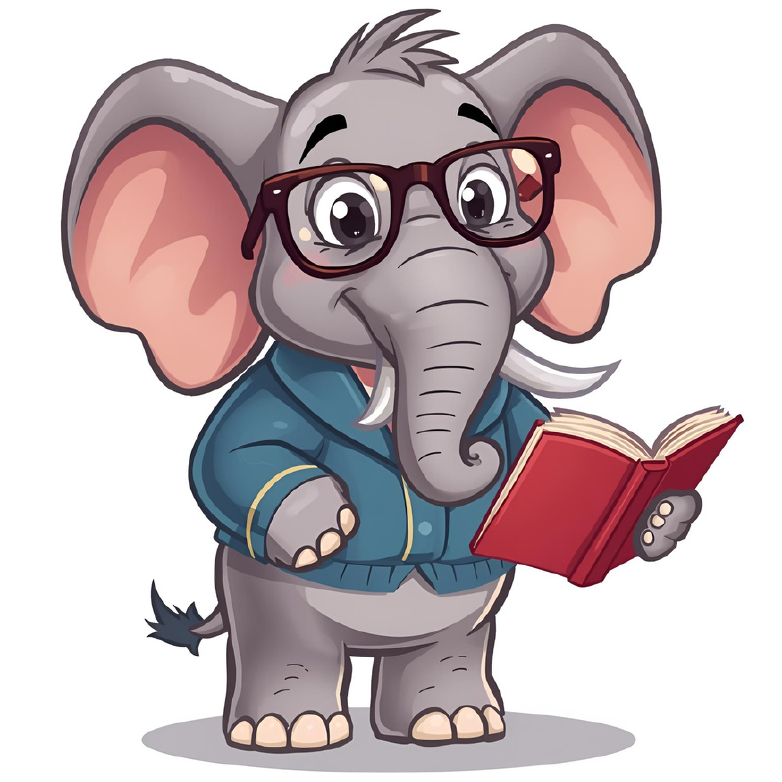}
}
\usepackage{booktabs}
\usepackage{makecell}
\usepackage{pifont}
\usepackage{xcolor}
\usepackage{colortbl}
\usepackage{tikz}

\title{\textsc{Eka-Eval} \emojismile : An Evaluation Framework for Low-Resource Multilingual Large Language Models}
\author{
\textbf{Samridhi Raj Sinha\textsuperscript{$\star$$\scriptscriptstyle\Diamond$}\thanks{Work done while interning at IIT Gandhinagar (SRIP).}}, 
\textbf{Rajvee Sheth\textsuperscript{$\S$$\scriptscriptstyle\Diamond$}},
\textbf{Abhishek Upperwal\textsuperscript{$\dagger$}},
\textbf{Mayank Singh\textsuperscript{$\S$$\scriptscriptstyle\Diamond$}}
\\
\\
 \textsuperscript{$\star$}NMIMS,
 \textsuperscript{$\dagger$}Soket AI,
 \textsuperscript{$\S$}Indian Institute of Technology Gandhinagar, 
\textsuperscript{$\scriptscriptstyle\Diamond$}LINGO Research Group
\\
 \small{
   \textbf{Correspondence:} \href{mailto:lingo@iitgn.ac.in}{singh.mayank@iitgn.ac.in}
 }
}

\begin{document}
\maketitle
\begin{abstract}
% The rapid advancement of Large Language Models (LLMs) has intensified the need for evaluation frameworks that address the requirements of linguistically diverse regions such as India and move beyond English-centric benchmarks. We introduce \textbf{\textsc{Eka-Eval}}, a unified framework that prioritizes usability through a dual-interface design (zero-code UI and interactive CLI) and integrates \textbf{55+ benchmarks} (Global and Indic) across \textbf{nine} major evaluation categories. The framework seamlessly supports both Hugging Face and proprietary models, offering \textbf{11 essential capabilities} through a modular, plug-and-play architecture. As the first end-to-end suite designed to enable scalable, multilingual LLM benchmarking across diverse domains, \textsc{Eka-Eval} combines modular workflows with dedicated support for low-resource Indian languages, enabling inclusive and comprehensive evaluation. Through extensive comparisons with five baselines, \textsc{Eka-Eval} demonstrates superior usability and flexibility, achieving the highest participant ratings, fastest setup time, and reliable benchmark reproducibility.

The rapid evolution of Large Language Models has underscored the need for evaluation frameworks that are globally applicable, flexible, and modular, and that support a wide range of tasks, model types, and linguistic settings. We introduce \textbf{\textsc{Eka-Eval}}, a unified, end-to-end framework that combines a \textbf{zero-code web interface} and an interactive CLI to ensure broad accessibility. It integrates \textbf{55+ diverse benchmarks} across \textbf{nine} evaluation categories, supports local and proprietary models, and provides \textbf{11 core capabilities} through a modular, plug-and-play architecture. Designed for scalable, multilingual evaluation with support for low-resource multilingual languages, \textsc{Eka-Eval} is, to the best of our knowledge, the \textbf{first suite} to offer comprehensive coverage in a single platform. Comparisons against \textbf{five existing baselines} indicate improvements of at least \textbf{2x better} on key usability measures, with the highest user satisfaction, faster setup times, and consistent benchmark reproducibility.

% The framework is open-source and publicly available at \url{https://github.com/lingo-iitgn/eka-eval} and a part of ongoing \textsc{Eka} initiative (\url{https://eka.soket.ai}), which aims to scale up to over 100 benchmarks and establish a robust, multilingual evaluation ecosystem for LLMs.

\begin{tabular}{@{}c@{\hspace{0.5em}}l@{\hspace{1em}}l@{}}
\raisebox{-0.1ex}{\faGlobe} & \textbf{Website} & \href{https://lingo.iitgn.ac.in/eka-eval/}{Eka-Eval} \\[0.3em]
\raisebox{-0.1ex}{\faLaptop} & \textbf{Demo} & \href{https://bit.ly/Eka-Eval}{bit.ly/Eka-Eval} \\[0.3em]
\raisebox{-0.1ex}{\faGithub} & \textbf{Code} & \href{https://github.com/lingo-iitgn/eka-eval}{github.com/lingo-iitgn/eka-eval}
\end{tabular}

\end{abstract}

\section{Introduction}
\label{sec:intro}
Large Language Models (LLMs) growing capabilities
continue to reshape NLP, enabling impressive
generalization across diverse tasks, including
instruction following, reasoning, summarization,
translation, and tool use. With the advent of general-purpose foundation models such as GPT-4~\citep{achiam2023gpt}, Claude~\citep{anthropic2024claude35sonnet}, Gemini~\citep{team2023gemini}, Deepseek~\citep{liu2024deepseek}, and Llama-3~\citep{touvron2023llama}, the focus of research has increasingly shifted from building task-specific models to systematically evaluating these powerful systems. Evaluation plays a critical role not only in measuring progress but in identifying capabilities, exposing limitations, and informing deployment strategies.

\begin{figure}[t]
    \centering
    \includegraphics[width=1.0\linewidth]{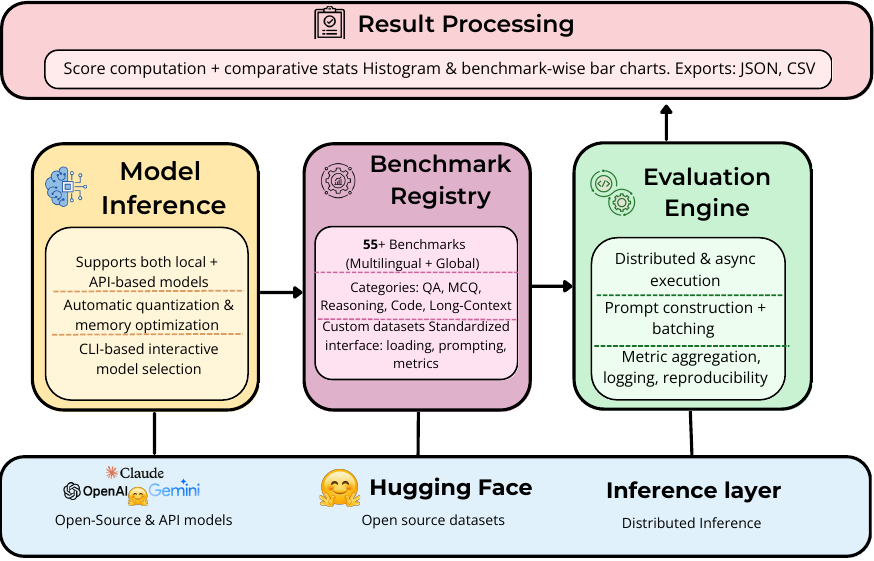}
    \caption{Architecture of \textsc{Eka-Eval}:  A modular framework combining model inference, benchmark registry, evaluation engine, and result processing with support for caching and distributed inference.}
    \vspace{-\baselineskip}
    \label{fig:architecture1}
\end{figure}

\begin{table*}[t]
    \centering
    \small
    \setlength{\tabcolsep}{1.7pt} 
    \renewcommand{\arraystretch}{1.25}
    
    \begin{tabular}{@{}l|cccc|cccc|ccc@{}}
        \toprule
        & \multicolumn{4}{c|}{\textbf{Core Flexibility}} 
        & \multicolumn{4}{c|}{\textbf{Advanced Capabilities}} 
        & \multicolumn{3}{c}{\textbf{Usability \& Specialization}} \\
        
        \textbf{Framework} 
        & \makecell{\textbf{Custom}\\\textbf{Data}} 
        & \makecell{\textbf{Custom}\\\textbf{Models}} 
        & \makecell{\textbf{Custom}\\\textbf{Prompt}} 
        & \makecell{\textbf{Quant-}\\\textbf{ization}}
        
        & \makecell{\textbf{Long}\\\textbf{Context}} 
        & \makecell{\textbf{Tool}\\\textbf{Use}} 
        & \makecell{\textbf{Dist.}\\\textbf{Infer.}} 
        & \makecell{\textbf{Visual}\\\textbf{Analysis}}
        
        & \makecell{\textbf{Zero-Code}\\\textbf{UI}} 
        & \makecell{\textbf{Interact.}\\\textbf{CLI}} 
        & \makecell{\textbf{Low-Resource}\\\textbf{Multilingual}} \\
        \midrule
        
        \texttt{lm-eval-harness}         & \cmark & \cmark & \cmark & \cmark & \cmark & \xmark & \cmark & \xmark & \xmark & \xmark & \xmark \\
        \texttt{OpenCompass}     & \cmark & \cmark & \cmark & \cmark & \cmark & \cmark & \cmark & \cmark & \cmark & \xmark & \xmark \\
        \texttt{HELM}            & \cmark & \cmark & \cmark & \cmark & \cmark & \cmark & \cmark & \cmark & \xmark & \xmark & \xmark \\
        \texttt{OpenAI Evals}    & \cmark & \cmark & \cmark & \xmark & \xmark & \xmark & \xmark & \cmark & \xmark & \xmark & \xmark \\
        \texttt{DeepEval}        & \cmark & \cmark & \cmark & \xmark & \cmark & \cmark & \xmark & \cmark & \xmark & \cmark & \xmark \\
        \texttt{FreeEval}        & \cmark & \cmark & \cmark & \cmark & \xmark & \xmark & \cmark & \cmark & \xmark & \cmark & \xmark \\
        \texttt{indic-eval}      & \xmark & \cmark & \xmark & \xmark & \xmark & \xmark & \cmark & \xmark & \xmark & \xmark & \cmark \\
        \texttt{UltraEval}       & \cmark & \cmark & \cmark & \cmark & \xmark & \xmark & \cmark & \cmark & \xmark & \xmark & \xmark \\
        \texttt{GlotEval}       & \cmark & \cmark & \cmark & \cmark & \xmark & \xmark & \xmark & \xmark & \xmark & \xmark & \cmark \\
        \midrule
        \textbf{\textsc{Eka-Eval}} & \cmark & \cmark & \cmark & \cmark & \cmark & \cmark & \cmark & \cmark & \cmark & \cmark & \cmark \\
        \bottomrule
    \end{tabular}
    
    \caption{Feature comparison with state-of-the-art frameworks. \textbf{\textit{Takeaway}}: \textsc{Eka-Eval} is the only framework to achieve full coverage across all eleven dimensions, uniquely combining a \textbf{zero-code UI} and \textbf{low-resource multilingual} support with advanced capabilities such as long-context and tool-use evaluation.}
    \label{tab:eval-comparison-full}
    \vspace{-0.3cm}
\end{table*}

\par \noindent \textbf{An Alarming Need}: Existing evaluation frameworks for LMs, such as \texttt{HELM}~\citep{liang2022holistic}, \texttt{lm-eval-harness}~\citep{gao2021framework}, \texttt{OpenCompass}~\citep{wu2023opencompass}, \texttt{DeepEval}~\citep{deepeval}, and \texttt{UltraEval}~\citep{he-etal-2024-ultraeval} predominantly target high-resource languages and rely on command-line interfaces with limited benchmark coverage~\citep{watts2024pariksha}. These frameworks create significant barriers for non-technical users who require graphical interfaces and researchers working with low-resource and multilingual settings. The challenge is especially pronounced in linguistically diverse regions globally, where numerous languages serve large populations; yet, existing evaluation frameworks lack comprehensive multilingual support and culturally grounded benchmarks, limiting rigorous assessment of language model capabilities across diverse linguistic settings \citep{pomerenke2025ai, xuan-etal-2025-mmlu, vayani2025all}.
%accessibility gap
% Furthermore, popular frameworks such as \texttt{OpenCompass} \citet{wu2023opencompass} and \texttt{HELM} ~\citep{liang2022holistic} require extensive configuration and engineering expertise, limiting their adoption by developers and researchers operating in low-resource environments. These challenges create a need for an actively maintained, community-driven, multilingual, and task-diverse evaluation suite.

% \par \noindent \textbf{Limitations}: 
% While specialized benchmarks such as IndicGLUE~\citep{kakwani2020indicnlpsuite}, IndicGenBench~\citep{singh2024indicgenbench}, and MILU~\citep{verma-etal-2025-milu} have advanced evaluation for Indian languages, they remain isolated initiatives lacking integration into unified pipelines. 

\par \noindent \textbf{Motivation}: Existing evaluation frameworks suffer from several limitations, such as \textit{(1)} absence of user-friendly interfaces, requiring extensive command-line expertise, broken-pipelines, and manual configuration \citep{gao2021framework, liang2022holistic},
\textit{(2)} restricted flexibility in benchmark selection and task customization \citep{wang2018glue, srivastava2022beyond},
\textit{(3)} insufficient task diversity, with fragmented support for advanced capabilities like long-context reasoning and tool use \citep{qin2023toolbench, zhang-etal-2024-bench}, and
\textit{(4)} limited multilingual coverage, particularly for low-resource languages \citep{ahuja-etal-2023-mega, luo-etal-2025-gloteval}.

\noindent To address this, we propose \textsc{Eka-Eval}, a uniquely scalable and extensible evaluation framework that unifies multilingual benchmarks under flexible task configurations. It enables seamless model integration and multi-platform accessibility, establishing a new benchmark for low-resource multilingual LLM evaluation frameworks.

% ADD something here:
% Two-liner USP about EKA

% It covers nine major evaluation categories: \textbf{\textit{(1)}} Code Generation and programming; \textbf{(2)} Mathematics and logical reasoning; \textbf{\textit{(3)}} Reading comprehension; \textbf{\textit{(4)}} Commonsense reasoning; \textbf{\textit{(5)}} World knowledge; \textbf{\textit{(6)}} Long-context understanding; \textbf{\textit{(7)}} General reasoning and knowledge; \textbf{\textit{(8)}} Tool use and API reasoning; and \textbf{\textit{(9)}} Indic-specific NLP benchmarks.

% including ARC-C~\citep{clark2018think}, BoolQ~\citep{clark2019boolq}, GSM8K~\citep{cobbe2021training}, Math~\citep{hendrycks2021measuring}, HellaSwag~\citep{zellers2019hellaswag}, MMLU~\citep{hendrycks2021mmlu}, SQuAD~\citep{rajpurkar2018know}, APIBench~\citep{patil2023api}, IndicGenBench~\citep{singh2024indicgenbench}.

\par \noindent \textbf{Contributions}: The key contributions are:
\begin{itemize}[nosep]

\item We propose \textsc{Eka-Eval}, a unified and modular evaluation framework that integrates \textbf{over 55 multilingual benchmarks} across nine major evaluation categories.

\item We introduce an open-source LLM evaluation 
framework that unifies a \textbf{zero-code UI}, an interactive CLI, and built-in support for low-resource multilingual settings, along with visualisations, and leaderboard, making large-scale evaluation accessible to both non-technical and technical users.

\item Through a comprehensive human evaluation involving eleven participants comparing \textbf{five existing frameworks}, showing that \textsc{Eka-Eval} achieves the highest participant ratings across all \textbf{six features}, performing at least \textbf{2× better} than existing frameworks in key usability and effectiveness.

%We provide extensive benchmark coverage spanning both English and Indic tasks, and conduct a structured human evaluation study with eleven participants across five existing frameworks on four benchmark tasks, demonstrating the effectiveness, usability, and reliability of \textsc{Eka-Eval}.

\end{itemize}

\section{Relevant Work} \label{sec:rel_work}

% The LLM evaluation landscape has steered from isolated benchmarks to modular, multi-dimensional frameworks, but gaps persist in multilingual coverage, Indic-language specialization, and usability.

\noindent \textbf{General-Purpose Frameworks:}
Early multi-task suites such as GLUE~\citep{wang2018glue} and SuperGLUE~\citep{wang2019superglue} established foundational standards. More comprehensive frameworks, including \texttt{HELM}~\citep{liang2022holistic}, \texttt{BIG-Bench}~\citep{srivastava2022beyond}, and \texttt{lm-eval-harness}~\citep{gao2021framework}, introduced extensible evaluation pipelines but typically require substantial technical effort and provide limited multilingual capabilities. Modular ecosystems such as \texttt{FreeEval}~\citep{yu2024freeeval} and \texttt{OpenCompass}~\citep{wu2023opencompass} offer advanced features like distributed inference. While \texttt{OpenCompass} covers extensive benchmarks and zero-code UI for limited models, custom evaluations still require CLI interaction or manual YAML configuration.

\noindent \textbf{Specialized Capability Benchmarks:}
Several domain-specific benchmarks target advanced LLM abilities, such as tool use in ToolBench~\citep{qin2023toolbench}, agentic behavior in API-Bank~\citep{li2023api} and InfiniteBench~\citep{zhang-etal-2024-bench}, long-context reasoning in RULER~\citep{hsieh2024ruler}, and mathematical and coding competencies in GSM8K~\citep{cobbe2021training} and HumanEval~\citep{chen2021evaluating}. While powerful, these benchmarks are fragmented across independent repositories, requiring users to maintain multiple codebases to conduct a thorough evaluation.

\noindent \textbf{Multilingual Evaluation:}
Multilingual suites such as XTREME~\citep{hu2020xtreme} and MEGA~\citep{ahuja-etal-2023-mega} cover 30+ languages but often rely on machine-translated content that lacks cultural grounding. Recent efforts of MCEval~\citep{Huang2025MCEvalAD} (13 cultures, 39K dynamic instances), GlotEval~\citep{luo-etal-2025-gloteval} (language-agnostic, culturally adaptive prompts), BLEnD~\citep{myung2024blend} (16 cultures, 52K native QA pairs), CulturalBench~\citep{chiu2024culturalbench} (45 global regions), and ALM-Bench~\citep{vayani2025all} (100 languages with cultural images) address this gap. However, these benchmarks still face insufficient data for low-resource languages, cultural biases favor dominant dialects, and evaluation metrics struggle to capture cultural appropriateness.

Existing frameworks are often complex and poorly modularized, hindering seamless incorporation into research workflows~\citep{he-etal-2024-ultraeval}. Furthermore, most toolkits also remain disproportionately focused on English and a handful of high-resource languages, making non-English-centric evaluations less practical for many researchers~\citep{luo-etal-2025-gloteval}, and none offer a zero-code UI or native support for low-resource languages, creating a gap for practitioners evaluating models in low-resource settings without dedicated systems expertise.

\noindent \textsc{Eka-Eval} addresses these gaps by unifying 55+ 
benchmarks across nine evaluation categories within a single 
modular workflow, with a strong focus on
low-resource multilingual evaluation. It is, to 
our knowledge, the only framework to pair a zero-code web UI 
along with interactive CLI, easy-to-use framework making rigorous evaluation accessible to non-technical users.

\section{Capabilities of LLM Evaluation Frameworks}
\label{sec:qual_comp}

We identify eleven foundational capabilities critical for modern LLM evaluation frameworks, informed by established systems like \texttt{lm-eval-harness}~\citep{gao2021framework}, \texttt{OpenCompass}~\citep{wu2023opencompass}, \texttt{FreeEval}~\citep{yu2024freeeval}, and \texttt{DeepEval}~\citep{deepeval}. These capabilities are organized into three categories: 

\noindent \textbf{\underline{A. Core Flexibility:}}\\
\textbf{\textit{(1) Custom Datasets:}} Support for user-defined evaluation data beyond standard benchmarks.\\
\textbf{\textit{(2) Custom Models:}} Compatibility with local checkpoints and API-hosted endpoints.\\
\textbf{\textit{(3) Custom Prompting:}} Flexible templates supporting zero-shot, few-shot, and chain-of-thought configurations.\\
\textbf{\textit{(4) Quantization:}} Support for memory-efficient quantized formats (8-bit, 4-bit weights).

\noindent \textbf{\underline{B. Advanced Capabilities:}}\\
\textbf{\textit{(5) Long Context:}} Processing extended inputs exceeding 4,000 tokens.\\
\textbf{\textit{(6) Tool Use:}} Evaluating agent-like behavior including API calls and multi-step reasoning.\\
\textbf{\textit{(7) Distributed Inference:}} Parallelized evaluation across multiple compute nodes.\\
\textbf{\textit{(8) Visual Analysis:}} Generation of interpretable visualizations (bar charts, plots, heatmaps, etc.).

\noindent \textbf{\underline{C. Usability \& Specialization:}}\\
\textbf{\textit{(9) Zero-Code UI:}} Web-based interface enabling end-to-end evaluation workflows without coding.\\
\textbf{\textit{(10) Interactive CLI:}} Command-line interface for configuring datasets, models, and strategies.\\
\textbf{\textit{(11) Low-Resource Multilingual:}} Comprehensive multilingual benchmarks featuring language-specific tokenization.\\
\begin{comment}

\end{comment}

\noindent Table~\ref{tab:eval-comparison-full} demonstrates that \textsc{Eka-Eval} achieves complete technical, linguistic, and usability coverage. While \texttt{lm-eval-harness} and \texttt{HELM} offer strong capabilities, they demand substantial programming expertise. \texttt{OpenCompass} includes a zero-code UI, but restricts it to leaderboard models and pre-configured benchmarks. \texttt{OpenAI Evals} and \texttt{DeepEval} support custom pipelines but lack quantization, long-context, and multilingual capabilities. Similarly, \texttt{UltraEval} and \texttt{GlotEval} provide provide robust customization and visual analysis (with the latter targeting low-resource languages) but but lack tool-use, long-context evaluation, 
and a zero-code UI. \texttt{indic-eval} is limited to seven benchmarks and offers no general-purpose capabilities. Critically, no 
existing framework pairs a zero-code UI with native low-resource multilingual support; \textsc{Eka-Eval} addresses both, lowering barriers to rigorous multilingual assessment.
% This comparison was cross-verified through participant feedback and official framework documentation, reinforcing \textsc{Eka-Eval} as a practical choice for global, Indic, and cross-lingual LLM evaluations.

\section{\textsc{Eka-Eval}}

\subsection{Design and Implementation}

\label{sec:arch}
\textsc{Eka-Eval} is architected as a modular, extensible evaluation framework that balances practical usability with comprehensive benchmark coverage. It provides a zero-code, web-based interface that allows users to run benchmarks, adjust parameters, and visualize results through interactive analytics and leaderboards, while a CLI enables detailed configuration and large-scale evaluations. The framework supports flexible model selection for both local and API-based models and is built around three core principles: \textit{modularity} for seamless extension and customization; \textit{accessibility} to support diverse environments, including low-resource language settings; and \textit{comprehensiveness} to cover a wide range of capabilities and low-resource multilingual benchmarks, with particular attention to underserved areas such as long-context reasoning and tool use.

\subsection{System Architecture}
The framework has a layered architecture with four main components (as illustrated in Figure~\ref{fig:architecture1}):
% the \textbf{Evaluation Engine}, \textbf{Benchmark Registry}, \textbf{Model Interface Layer}, and \textbf{Results Processing System}, 

% each containing specialized secondary components to handle specific functionality. This hierarchical design enables seamless integration of new benchmarks, model backends, and evaluation metrics while maintaining backwards compatibility and reproducibility (see Figure~\ref{fig:architecture1}).

\subsubsection{Evaluation Engine}
Manages all evaluation workflows:

\noindent \textbf{Task Scheduler:} Manages task scheduling, prompt formatting, and result aggregation across distributed inference setups. The scheduler implements intelligent work distribution, as shown in \texttt{main\_orchestrator()}, by dynamically assigning evaluation tasks to available workers based on resource constraints and model requirements.

\noindent \textbf{Batch Optimizer:} Implements intelligent batching strategies and supports various quantization schemes to optimize memory usage and inference speed. As seen in the PIQA
evaluation implementation, the optimizer automatically adjusts \texttt{generation\_batch\_size} parameters to maximize throughput while preventing out-of-memory errors.

\noindent \textbf{Distributed Coordinator:} Coordinates evaluation across multiple GPUs and workers using Python's multiprocessing library, launching multiple \texttt{worker\_process} instances to execute independent evaluation tasks in parallel across benchmarks and model configurations.

\subsubsection{Benchmark Registry}
Provides a unified interface for managing datasets:

\noindent \textbf{Dataset Manager:} The \texttt{BenchmarkRegistry} class handles diverse dataset formats and sources, abstracting the complexities of different evaluation protocols. The manager supports datasets from HuggingFace Hub, local files, and custom API-models through standardized interfaces.

\subsubsection{Model Interface Layer}

Access to different local and API-based models:

% The Model Interface Layer provides unified abstraction across local and API-based models through four secondary components:

\noindent \textbf{Local Model Loader:} 
Initializes transformer-based checkpoints with automatic device allocation and quantization.
% Initializes models using HuggingFace transformers with automatic device allocation, quantization support, and memory optimization. The loader handles diverse model formats while maintaining consistent interfaces.

\noindent \textbf{API Client Manager:} Manages proprietary endpoints through dedicated clients (\texttt{OpenAIClient}, \texttt{GeminiClient}, \texttt{ClaudeClient}) that extend \texttt{BaseAPIClient}, providing unified request handling with rate limiting and authentication.

\noindent \textbf{Interactive Selection Interface:} Implements \texttt{get\_model\_selection\_interface()} for dynamic model discovery and selection, supporting local model paths and API configurations(See Appendix~\ref{sec:appendix-cli}). 
% with environment variable management.

\noindent \textbf{Resource Manager:} Ensures efficient memory management via cleanup functions, preventing resource leaks during repeated evaluation runs.

\begin{table*}[t]
    \centering
    \small
    \setlength{\tabcolsep}{4.5pt}
    \begin{tabular}{lcccccc}
      \toprule
      \textit{\textbf{Framework}} & \textit{\textbf{Setup} \& \textbf{Config}} & 
      \textit{\textbf{Navigation}} & \textit{\textbf{UI}} & 
      \textit{\textbf{Result Export}} & \textit{\textbf{Extensibility}} & 
      \textit{\textbf{Multilingual Support}} \\ 
      \toprule

      \texttt{lm-eval-harness} 
      & 3.73 $\pm$ 0.72 & 3.91 $\pm$ 0.66 & 1.00 $\pm$ 0.00 
      & 3.64 $\pm$ 0.70 & 4.18 $\pm$ 1.00 & 1.64 $\pm$ 0.50 \\

      \texttt{OpenCompass} 
      & 2.18 $\pm$ 0.93 & 2.36 $\pm$ 0.67 & 3.09 $\pm$ 1.36 
      & 3.09 $\pm$ 0.93 & 2.80 $\pm$ 1.06 & 2.09 $\pm$ 0.71 \\

      \texttt{HELM} 
      & 1.91 $\pm$ 0.71 & 2.64 $\pm$ 1.05 & 2.36 $\pm$ 1.22 
      & 2.55 $\pm$ 1.42 & 2.09 $\pm$ 0.93 & 1.36 $\pm$ 0.50 \\

      \texttt{indic-eval} 
      & 2.55 $\pm$ 0.88 & 3.18 $\pm$ 0.67 & 1.00 $\pm$ 0.00 
      & 2.55 $\pm$ 1.00 & 2.55 $\pm$ 0.88 & 4.55 $\pm$ 0.53 \\

      \texttt{FreeEval} 
      & 2.55 $\pm$ 1.24 & 2.45 $\pm$ 0.71 & 2.73 $\pm$ 1.01 
      & 2.64 $\pm$ 1.24 & 2.36 $\pm$ 0.87 & 1.45 $\pm$ 0.53 \\ \hline

      \textbf{\textsc{Eka-Eval}} 
      & \textbf{4.27 $\pm$ 0.93} & \textbf{4.55 $\pm$ 0.73} 
      & \textbf{4.64 $\pm$ 0.53} & \textbf{4.55 $\pm$ 0.50} 
      & \textbf{4.64 $\pm$ 0.50} & \textbf{4.73 $\pm$ 0.50} \\ 

      \bottomrule
    \end{tabular}
    \caption{Average participant ratings of the evaluation frameworks by eleven participants (mean $\pm$ standard deviation; Likert scale: 1–5, 1 = poor and 5 = excellent). \textbf{\textit{Takeaway}}: \textsc{Eka-Eval} achieves the highest mean ratings across all six dimensions, with notably stronger scores on Zero-Code UI and Global Low-Resource Language Support.}
    \label{tab:framework-results-comparison}
\end{table*}

\begin{table}[t]
    \centering
    \small
    \setlength{\tabcolsep}{25pt}
    \begin{tabular}{lc}
      \toprule
      \textit{\textbf{Framework}} & \textit{\textbf{Time Taken}} \\ 
      \toprule
      \texttt{lm-eval-harness} & $22 \pm 7.58$ \\
      \texttt{OpenCompass} & $36 \pm 9.83$ \\
      \texttt{HELM} & $58 \pm 5.70$ \\
      \texttt{IndicEval} & $32 \pm 4.14$ \\
      \texttt{FreeEval} & $35 \pm 7.07$ \\
      \hline
      \textbf{\textsc{Eka-Eval}} & $\mathbf{11 \pm 3.18}$ \\
      \bottomrule
    \end{tabular}
    \caption{Comparison of time taken for installation and configuration across six frameworks reported across five successful runs on GPU-setup. \textbf{\textit{Takeaway}}: \textsc{Eka-Eval} records the lowest setup time and variance.}
    \vspace{-\baselineskip}
    \label{tab:framework_time_comparison}
\end{table}

% \begin{table*}[t]
%     \centering
%     % \scriptsize
%     \setlength{\tabcolsep}{6pt}
%     \begin{tabular}{lccccc}
%       \toprule
%       \textit{\textbf{Framework}} & \textit{\textbf{Setup} \& \textbf{Config}} & \textit{\textbf{Navigation}} & \textit{\textbf{Result Export}} & \textit{\textbf{Indic Support}} & \textit{\textbf{Extensibility}} \\ \toprule
%       \texttt{lm-eval-harness} & 3.67 $\pm$ 0.58 & 4.00 $\pm$ 1.00 & 4.00 $\pm$ 1.73 & 3.33 $\pm$ 2.08 & 4.33 $\pm$ 1.15 \\
%       \texttt{OpenCompass} & 3.33 $\pm$ 0.58 & 3.33 $\pm$ 0.58 & 3.67 $\pm$ 0.58 & 3.00 $\pm$ 1.00 & 3.67 $\pm$ 0.58 \\
%       \texttt{HELM} & 3.33 $\pm$ 0.58 & 3.67 $\pm$ 0.58 & 4.00 $\pm$ 1.00 & 2.33 $\pm$ 0.58 & 3.67 $\pm$ 0.58 \\
%       \texttt{indic-eval} & 3.67 $\pm$ 0.58 & 3.67 $\pm$ 1.15 & 4.00 $\pm$ 1.00 & 3.67 $\pm$ 0.58 & 3.00 $\pm$ 1.00 \\
%       \texttt{FreeEval} & 3.00 $\pm$ 1.00 & 2.67 $\pm$ 1.15 & 4.33 $\pm$ 0.58 & 2.67 $\pm$ 1.53 & 3.00 $\pm$ 1.00 \\ \hline
%       \textbf{\textsc{Eka-Eval}} & 3.67 $\pm$ 0.58 & \textbf{4.67 $\pm$ 0.58} & \textbf{4.67 $\pm$ 0.58} & \textbf{5.00 $\pm$ 0.00} & \textbf{4.67 $\pm$ 0.58} \\ \bottomrule
%     \end{tabular}
%     \caption{Average participant ratings of the evaluation frameworks by three participants (mean $\pm$ standard deviation; Likert scale: 1–5). \textsc{Eka-Eval} achieves the highest ratings in four functionalities compared to existing frameworks.}
%     \label{tab:framework-results-comparison}
% \end{table*}

\subsubsection{Results Processing System}
Handles comprehensive output management through three secondary components:

\noindent \textbf{Metrics Calculator:} Employs HuggingFace's industry-standard \texttt{evaluate} library to compute essential metrics such as accuracy, BLEU \citep{papineni2002bleu}, F1-score~\citep{10.1145/3606367}, exact match~\citep{rajpurkar2016squad}, and Pass@1~\citep{chen2021evaluating}, ensuring consistency with widely adopted benchmarks and standard evaluation practices across diverse LLM tasks.

% It also implements robust error handling for edge cases and missing data; for example, when a model returns ``The answer is probably B'', regex-based extraction retrieves the label; if that fails, a default score is assigned.

\noindent \textbf{Visualisations analytics:} Provides comparative analysis across multiple models and benchmark configurations by generating visualizations such as bar charts, heatmaps, and radar plots (including support for cross-model comparisons).

\noindent \textbf{Export Manager:} Handles result export in multiple formats, including JSON and CSV. The manager maintains evaluation metadata including model parameters, benchmark versions, execution timestamps, and system configurations.
\subsubsection{Interface and Deployment Layer}
Handles complex evaluation logic with ease of use through a streamlined full-stack user interface (See Appendix \S\ref{sec:zero-code-appendix} and Figure \ref{fig:arch}).

\noindent \textbf{Full-Stack Architecture:} A decoupled \textsc{React} \cite{fedosejev2015react} frontend and \textsc{FastAPI} \cite{tiangolo2018fastapi} backend, served through \textsc{Nginx}, provide a stable and scalable deployment layer.

\noindent \textbf{Real-Time Telemetry:} WebSocket-based streaming delivers live inference logs and GPU status directly to the UI, eliminating the need for CLI.

\noindent \textbf{LLM-Powered Diagnostics:} An LLM (LLaMA-3.3 70B~\citep{grattafiori2024llama}) interprets evaluation outputs and generates summaries of model performance and failure patterns, revealing additional insights.

\noindent \textbf{Interactive Dashboard:} A ``\textit{Zero-Code}'' user-interface with visual comparisons and live leaderboards enables users to run and analyze evaluations without technical expertise.

%The \textbf{Results Processing System} handles metric computation, statistical analysis, and report generation. It supports rich analytics including confidence intervals and comparative analysis across multiple models and benchmark configurations. Results are exported in multiple formats (JSON, CSV) to support downstream analysis and visualization workflows.
\subsection{Comprehensive Benchmark Coverage}
\label{sec:benchmarks}

\textsc{Eka-Eval} covers nine major evaluation categories with comprehensive benchmark support across 55+ benchmarks (See Appendix~\ref{sec:appendix}). \\
These categories include:

\begin{enumerate}[noitemsep,nolistsep,leftmargin=*]
\item \textit{Code Generation and Programming:} These are assessed using HumanEval~\citep{chen2021evaluating}, MBPP~\citep{austin2021program}, HumanEval+~\citep{liu2023your}, PythonSaga~\citep{yadav-etal-2024-pythonsaga} and MBPP EvalPlus~\citep{liu2023your} with Pass@1 metrics.

\item \textit{Mathematics and Logical Reasoning:} Mathematical capabilities are evaluated through GSM8K~\citep{cobbe2021training}, MATH~\citep{hendrycks2021measuring}, and ARC-Challenge~\citep{clark2018think}.

\item \textit{Reading Comprehension:} Text understanding is evaluated using SQuAD~\citep{rajpurkar2018know}, QuAC~\citep{choi-etal-2018-quac}, and BoolQ~\citep{clark2019boolq} with F1 and exact match metrics.

\item \textit{Commonsense Reasoning:} We incorporate PIQA~\citep{bisk2020piqa}, SIQA~\citep{sap2019social}, HellaSwag~\citep{zellers2019hellaswag}, ARC-Easy/Challenge~\citep{clark2018think}, WinoGrande~\citep{sakaguchi2021winogrande}, CommonSenseQA~\citep{talmor2018commonsenseqa}, and OpenBookQA~\citep{mihaylov2018can}.

\item \textit{World Knowledge:} Factual knowledge is tested through TriviaQA~\citep{joshi2017triviaqa} and NaturalQuestions~\citep{kwiatkowski2019natural}.

\item \textit{Long-Context Understanding:} For extended context, we include ZeroSCROLLS~\citep{shaham2023zeroscrolls}, Needle-in-a-Haystack~\citep{wang-etal-2025-multimodal}, and InfiniteBench~\citep{zhang-etal-2024-bench}.

\item \textit{General Reasoning:} Foundational capabilities are tested via MMLU~\citep{hendrycks2021mmlu}, MMLU-Pro~\citep{wang2024mmlu}, IFEval~\citep{zhou2023instruction}, BBH~\citep{suzgun2022challenging}, and AGI-Eval~\citep{zhong-etal-2024-agieval}.

\item \textit{Tool Use and API Reasoning:} Practical capabilities are assessed through API-Bank~\citep{li2023api} and API-Bench~\citep{patil2023api}.

\item
\textit{Multilingual and Low-Resource Language Support:} LLMs are known to underperform on languages spoken across South Asia, Africa, and Southeast Asia, yet no existing evaluation framework provides sufficient coverage to diagnose this gap~\citep{luo-etal-2025-gloteval, kakwani2020indicnlpsuite}. \textsc{Eka-Eval} features the largest unified Multilingual suite (23 benchmarks), including:
\textbf{\textit{(i) Knowledge:}} IndicMMLU-Pro~\citep{sankalp2025indicmmlupro}, 
MMLU-IN~\citep{hendrycks2021measuring}, TriviaQA-IN~\citep{joshi2017triviaqa}, 
MILU~\citep{verma-etal-2025-milu} spanning Kannada, Odia, Malayalam, and Urdu. \textbf{\textit{(ii) Reasoning:}} HellaSwag-IN~\citep{zellers2019hellaswag}, 
ARC-C-IN~\citep{clark2018think}, IndicCOPA~\citep{kakwani2020indicnlpsuite}, 
XCOPA~\citep{ponti-etal-2020-xcopa} in Quechua, Indonesian, and Swahili, 
and GSM8K-IN~\citep{cobbe2021training} for reasoning across 
Indic languages. \textbf{\textit{(iii) Reading \& QA:}} Belebele~\citep{bandarkar-etal-2024-belebele} 
covering 122 languages including Yoruba, Oromo, and Indonesian; 
BoolQ-IN~\citep{clark2019boolq}, XQuAD-IN~\citep{artetxe2020cross} 
in Hindi and Greek, XorQA-IN~\citep{asai2021xor} in Bengali and Telugu, 
and Indic-QA~\citep{singh2025indicqabenchmarkmultilingual}. \textbf{\textit{(iv) Generation (NLG):}} Flores-IN~\citep{goyal-etal-2022-flores} for translation across 200 languages; IndicParaphrase, IndicWikiBio, IndicQuestionGeneration, IndicSentenceSummarization, and 
IndicHeadlineGeneration~\citep{singh2024indicgenbench} in Odia, 
Assamese and Punjabi. \textbf{\textit{(v) NLU:}} IndicNER and IndicSentiment in 11 languages, IndicGLUE~\citep{kakwani2020indicnlpsuite}, and 
XNLI~\citep{conneau-etal-2018-xnli} for cross-lingual inference 
in Arabic, Swahili, and Urdu.
    %  \textbf{\textit{(i) Knowledge:}} IndicMMLU-Pro~\citep{sankalp2025indicmmlupro}, MMLU-IN~\citep{hendrycks2021measuring}, TriviaQA-IN~\citep{joshi2017triviaqa}, MILU~\citep{verma-etal-2025-milu}.
    % \textbf{\textit{(ii) Reasoning:}} HellaSwag-IN~\citep{zellers2019hellaswag}, ARC-C IN~\citep{clark2018think}, IndicCOPA~\citep{kakwani2020indicnlpsuite}, XCOPA~\citep{ponti-etal-2020-xcopa} in Arabic, Indonesian, and Swahili, GSM8K-IN~\citep{cobbe2021training}. \textbf{\textit{(iii) Reading \& QA:}} Belebele \citep{bandarkar-etal-2024-belebele} covering 122 languages including Yoruba, Oromo, and Indonesian, BoolQ-IN~\citep{clark2019boolq}, XQuAD-IN~\citep{artetxe2020cross}, XorQA-IN~\citep{asai2021xor}, Indic-QA~\citep{singh2025indicqabenchmarkmultilingual}.
    % \textbf{\textit{(iv) Generation (NLG):}} Flores-IN~\citep{goyal-etal-2022-flores}, IndicParaphrase, IndicWikiBio, IndicQuestionGeneration, IndicSentenceSummarization, IndicHeadlineGeneration~\citep{singh2024indicgenbench}.
    % \textbf{\textit{(v) NLU:}} IndicNER, IndicSentiment, IndicGLUE~\citep{kakwani2020indicnlpsuite}, XNLI \citep{conneau-etal-2018-xnli} for in Arabic, Swahili, and Urdu.
\end{enumerate}

% \textsc{Eka-Eval} prioritizes extensibility through a low-code plugin architecture that allows users to integrate new benchmarks and custom metrics with minimal changes. Complex configurations ranging from parameter sweeps to prompt variations are managed via a hierarchical JSON system. Crucially, the user interface exposes these capabilities interactively, enabling researchers to modify inference parameters (e.g., temperature, batch size) and strategies in real-time without direct management of underlying configuration files.

\section{Experiments}

To assess the effectiveness and usability of \textsc{Eka-Eval}, we conducted a comprehensive evaluation combining benchmark coverage analysis and user-centered feedback across six prominent frameworks: \texttt{lm-eval-harness}, \texttt{OpenCompass}, \texttt{HELM}, \texttt{FreeEval}, \texttt{indic-eval}, and \textbf{\textsc{Eka-Eval}}. These frameworks were selected as representative state-of-the-art baselines spanning across widely adopted general-purpose to specialized multilingual evaluation suites, ensuring a rigorous comparison.

\paragraph{Evaluation Procedure}

% Eleven graduate-level computer science researchers with demonstrated proficiency in Git/GitHub evaluated each framework in a single-blind setup. 
Eleven graduate-level computer science researchers were selected from NLP and software engineering programs. Participants were stratified by prior experience using at least one existing evaluation framework (5 with, 6 without) and evaluated each framework in a single-blind setup. Selection was not tied to knowledge of \textsc{Eka-Eval}, mitigating demand bias; sample size was constrained by available GPU resources. Participants followed a standardized evaluation protocol in a controlled environment. They installed each framework, integrated HuggingFace models - including \texttt{gemma-2-2b}~\citep{team2024gemma} and \texttt{sarvam-1B}~\cite{sarvamai_sarvam1_2024}, and executed the recommended workflows on four diverse benchmarks: \textbf{WinoGrande}, \textbf{PIQA}, \textbf{ARC-C-IN} and \textbf{MMLU-IN}. 
% All participants had prior experience with LLM evaluation frameworks and were trained on the standardized protocol before the study. 

\noindent \noindent All ratings were recorded using the standardized protocol (Appendix \S\ref{sec:guidelines}) to ensure direct and consistent comparison across frameworks. Participants assessed each framework across six evaluation criteria and rated them using a Likert scale ranging from 1 to 5:

\begin{enumerate}[noitemsep,nolistsep,leftmargin=*]
    \item \textbf{Setup and Configuration Time:} 
    Time and effort required to install dependencies, configure models, and run an initial benchmark. 
    
    \item \textbf{Ease of Navigation:}
    Intuitiveness of navigation, benchmark selection, and configuration, including CLI clarity, documentation quality, and ease of discovering options.

     \item \textbf{Zero-Code UI Availability:}
    Presence and usability of GUI-based tools for visualization, benchmark execution, and result analysis, eliminating the need for command-line interaction.
    
    \item \textbf{Result Reporting and Export:}
    Clarity and accessibility of evaluation outputs, with options to export results (e.g., JSON, CSV) and create visualizations such as bar charts or heatmaps.
    
    \item \textbf{Extensibility:}
    Ease of customizing the framework to add new prompt templates, models, benchmarks, or evaluation metrics. 

    \item \textbf{Multilingual Language Support:}
    Support for multilingual benchmarks such as Belebele, ARC-IN, FLORES, and XNLI. 
\end{enumerate}

\noindent Table~\ref{tab:framework-results-comparison} presents average participant ratings across six evaluation criteria using a 1-5 Likert scale (1= poor or absence of functionality, 5= excellent or flexible capability). \textsc{Eka-Eval} achieved the highest ratings across all criteria, establishing superior usability and capability. Moreover, Table~\ref{tab:framework_time_comparison} summarizes the average setup times across six frameworks, which varied widely due to dependency issues and unstable codebases in several baselines. Nine of the eleven participants reported that \texttt{HELM} suffered from poor documentation and a broken pipeline, while \texttt{OpenCompass} required manual code edits and \texttt{indic-eval} experienced installation difficulties. \textsc{Eka-Eval} achieved the shortest setup time. Table~4 (See Appendix~\ref{tab:eval-benchmark-scores} for full results) presents reproduced benchmark scores on \texttt{gemma-2-2b}, showing how they align with broader framework differences and highlighting \textsc{Eka-Eval}'s consistency across global-multilingual benchmarks. \texttt{HELM} was excluded from this comparison, as it neither supports these benchmarks nor produced functional results despite repeated attempts.

\section{Conclusion and Future Work}
\label{sec:conc}

% In this work, we introduced \textsc{Eka-Eval}, a unified and extensible framework for the evaluation of LLMs across both global and Indic tasks. By supporting 35+ benchmarks and accommodating multiple model backends—including local checkpoints and API-based deployments—\textsc{Eka-Eval} enables reproducible, scalable, and backend-agnostic evaluation pipelines. 

We present \textsc{Eka-Eval}, an open-source unified evaluation framework designed to overcome persistent challenges in accessibility, reproducibility, and low-resource multilingual coverage for language model assessment. Our study with eleven participants demonstrates substantial usability improvements, achieving lowest setup time and the highest satisfaction ratings. \textsc{Eka-Eval} enables an interactive UI, rigorous, reproducible evaluation without manual coding expertise. The current release is an ongoing work, with 30 stars and 3 forks on GitHub. Future work will expand coverage to multimodal LLMs, benchmarks, and AI diagnostic capabilities.

\newpage

\section{Limitations}

\paragraph{Multimodal Support.} \textsc{Eka-Eval} currently 
supports text-only LLMs; multimodal model evaluation is 
left to future work.

\paragraph{Reproducibility.} Evaluation results may be 
affected by changes in external dataset or model versions; 
we recommend explicit versioning and caching for 
fully reproducible runs.

\paragraph{User study scope.} The usability study (n=11) constitutes a formative evaluation conducted under GPU resource constraints, providing initial usability insights and motivating broader future evaluations.

\paragraph{Inference Backend.} \textsc{Eka-Eval} currently relies on HuggingFace Transformers for model inference; vLLM backend integration is underway to enable faster, high-throughput evaluation at scale.

% \paragraph{Multilingual empirical coverage.} While \textsc{Eka-Eval} integrates 23 multilingual benchmarks, the empirical results in Table~4 cover only four benchmarks. Broader multilingual evaluation across additional models and language families remains future work.

\paragraph{Scalability.} The current release processes requests sequentially; supporting concurrent multi-user deployments may require additional infrastructure and compute beyond the scope of this work.

\section{Ethics Statement}
This research uses publicly available data without personally identifiable information. All datasets and models comply with their terms of use. The work is intended for academic research. Potential misuse or unintended amplification of biases should be carefully considered before deployment.

\section{Acknowledgements}

The authors express their gratitude to Himanshu Beniwal for helping with writing the manuscript and to Birudugadda Srivibhav, Sailesh Panda, Gautham Bharati, Indrayudh Mandal, Krudant Randai, Sahil Gawande, Tirth Bhatt, Shruti Bhat, Mann agarwal, Himanshu Sharma and Khushbu Bijawat for their contributions in evaluating the framework, reviewing the manuscript, and reporting the results. We also appreciate the valuable suggestions and feedback provided by Aamod Thakur, Prathamesh Shanbhag and Mahavir Patil.

\bibliography{custom}
\newpage
\appendix
\section{Appendix}
\label{sec:appendix}

\begin{table*}[t]
    \centering
    \small
    \setlength{\tabcolsep}{22pt}
    \begin{tabular}{lcccc}
      \toprule
      \textit{\textbf{Framework}} 
      & \textit{\textbf{Winogrande}} 
      & \textit{\textbf{PIQA}} 
      & \textit{\textbf{ARC-IN}} 
      & \textit{\textbf{MMLU-IN}} \\
      \toprule

      \texttt{lm-eval-harness} 
      & 65.8 $\pm$ 6.0  & 74.8 $\pm$ 3.5 & --  & -- \\

      \texttt{OpenCompass} 
      & 61.1 $\pm$ 5.0  & 73.3 $\pm$ 4.3   & --  & -- \\

      \texttt{indic-eval} 
      & 65.8 $\pm$ 1.9   & 78.7 $\pm$ 1.4  & 28.1 $\pm$ 2.3 
      & 35.0 $\pm$ 2.1 \\

      \texttt{FreeEval} 
      & 60.0 $\pm$ 4.5  & 54.5 $\pm$ 3.2  & --
      & -- \\ \hline

      \textbf{\textsc{Eka-Eval}} 
      & \textbf{66.5 $\pm$ 2.5}  & \textbf{77.5 $\pm$ 1.5} 
      & \textbf{31.5 $\pm$ 3.5}  & \textbf{33.5 $\pm$ 0.5} \\

      \bottomrule
    \end{tabular}

    \caption{Benchmark Score Parity. Comparison of benchmark scores (mean $\pm$ standard deviation). `--' represents those benchmarks are not supported for the frameworks. \textbf{\textit{Takeaway}}: \textsc{Eka-Eval} reproduces verified consistent scores with established frameworks and closely aligns with their actual benchmark results.} 
    \vspace{-\baselineskip}
    \label{tab:eval-benchmark-scores}
\end{table*}

\subsection{Multilingual benchmark configuration}
As demonstrated in Figure~\ref{fig:arc-challenge-config}, a sample configuration used to evaluate the ARC-Challenge-Indic benchmark across 11 Multilingual languages. It illustrates how task parameters, templates, and dataset references are modularly specified in \textsc{Eka-Eval}.

\vspace{0.5cm}
\begin{figure}[h]
\centering
\begin{minipage}{0.9\textwidth}
\begin{Verbatim}[fontsize=\footnotesize]
"ARC-Challenge-Indic": {
  "description": "Zero-shot evaluation 
    across 11 Indic languages",
  "evaluation_function": 
    "indic.arc_c_in.evaluate_arc_c_in",
  "task_args": {
    "dataset_name": 
      "sarvamai/arc-challenge-indic",
    "target_languages": ["bn", "en", 
      "gu", "hi", "kn", "ml", "mr", 
      "or", "pa", "ta", "te"],
    "dataset_split": "validation",
    "num_few_shot": 0,
    "max_new_tokens": 10,
    "generation_batch_size": 8,
    "prompt_template_name_zeroshot": 
      "arc_c_in_0shot",
    "prompt_template_name_fewshot": 
      "arc_c_in_5shot",
    "prompt_file_benchmark_key": 
      "arc_c_in",
    "prompt_file_category": "indic"
  }
}
\end{Verbatim}
\end{minipage}
\caption{ARC-Challenge-Indic benchmark configuration example}
\label{fig:arc-challenge-config}
\end{figure}

\subsection{Prompt Template System}

A critical component of \textsc{Eka-Eval} is its sophisticated prompt management system, which handles diverse evaluation paradigms and languages. The framework implements a flexible template system demonstrated through PIQA benchmark prompt \ref{fig:piqa-prompts}:

\begin{figure}[h!]
\centering
\begin{minipage}{0.48\textwidth}
\footnotesize
\begin{verbatim}
{
  "piqa_generation": {
    "template": "Choose the most appropriate 
      solution (0 or 1) to achieve the goal:
      \n\nQuestion: {goal}\n0) {sol1}
      \n1) {sol2}\nAnswer:",
    "description": "Generation-based PIQA prompt"
  },
  "piqa_5shot_generation": {
    "template_prefix": "Choose the most 
    "few_shot_example_template": 
      "Question: {goal}\n0) {sol1}
      \n1) {sol2}\nAnswer: {answer_label}",
    "few_shot_separator": "\n\n",
    "template_suffix": "Question: {goal}
      \n0) {sol1}\n1) {sol2}\nAnswer:",
    "description": "Few-shot generation prompt"
  },
  "default_few_shot_examples_piqa": [
    {
      "goal": "To remove a stain from clothing",
      "sol1": "Apply cold water immediately...",
      "sol2": "Set the clothing on fire...",
      "answer_label": "0"
    }
  ]
}
\end{verbatim}
\end{minipage}
\caption{PIQA prompt templates supporting multiple evaluation paradigms}
\label{fig:piqa-prompts}
\end{figure}
\noindent The Prompt template system as shown in Figure~\ref{fig:piqa-prompts} supports zero-shot, few-shot, and chain-of-thought prompting strategies, ensuring consistency across evaluation modes and languages. Users can customize prompt strategies and easily configure them in the benchmark configuration file, as shown in Figure 2.

\begin{figure}
        \centering
        \includegraphics[width=0.8\linewidth]{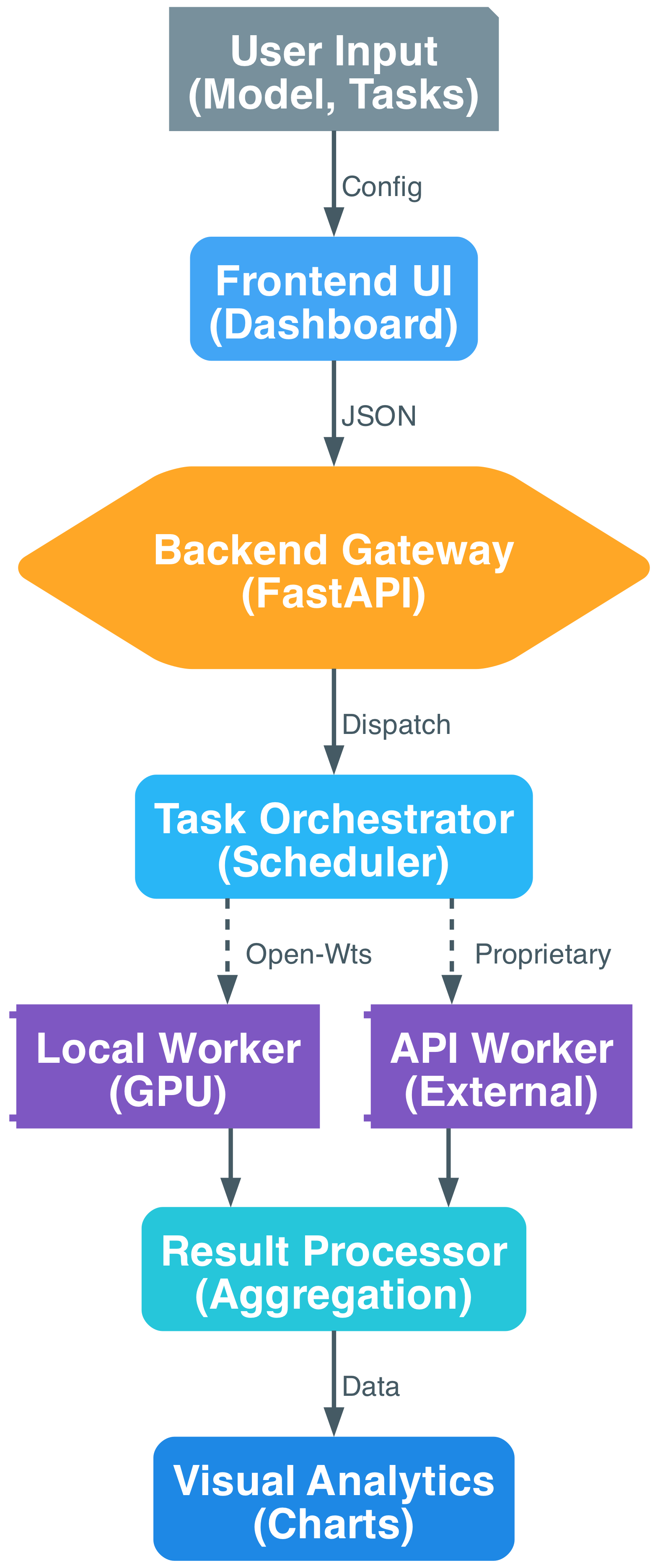}
        \caption{The overall architectural pipeline of the \textsc{Eka-Eval} framework.}
        \label{fig:arch}
\end{figure}

\section{Zero-code User Interface}
\label{sec:zero-code-appendix}

To democratize access for LLM evaluation, \textsc{Eka-Eval} includes a ``Zero-code'' user interface that mirrors the full capabilities of the CLI while adding advanced features for prompt configuration, resource management, and AI-assisted analysis, making rigorous evaluation accessible to both non-technical and technical users. Figure~\ref{fig:arch} illustrates the end-to-end architecture, spanning user input, model inference, evaluation execution, and result visualization across \textsc{Eka-Eval}'s four core components.

\begin{figure*}[b]
    \centering
    \includegraphics[width=1\linewidth]{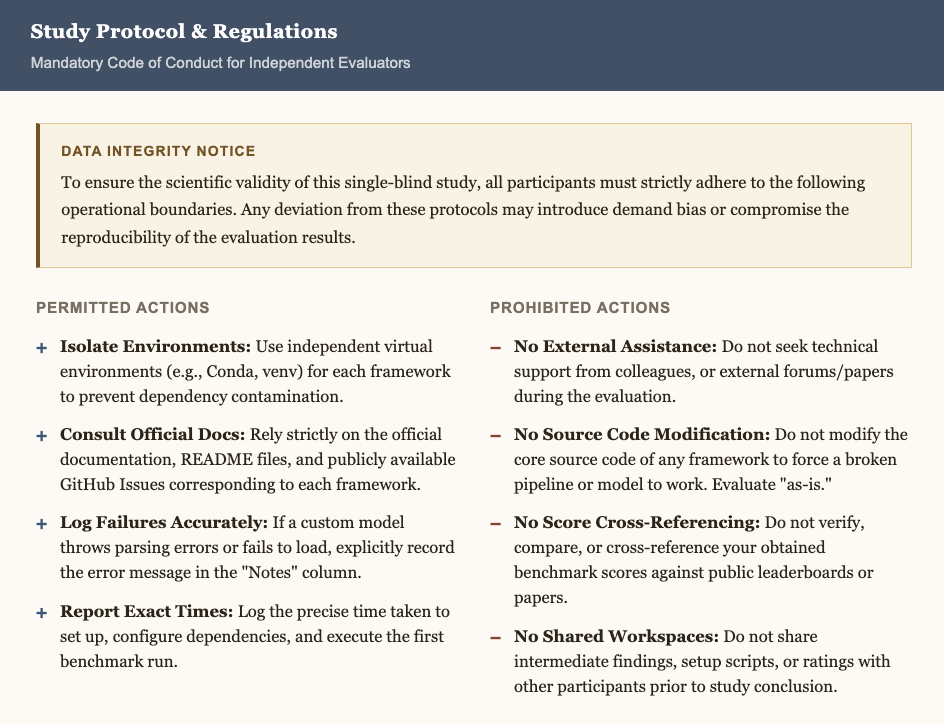}
    \caption{Standardized study protocol specifying permitted and prohibited actions for all participants.}
    \label{fig:rating}
\end{figure*}

\begin{figure*}[b]
    \centering
    \includegraphics[width=1\linewidth]{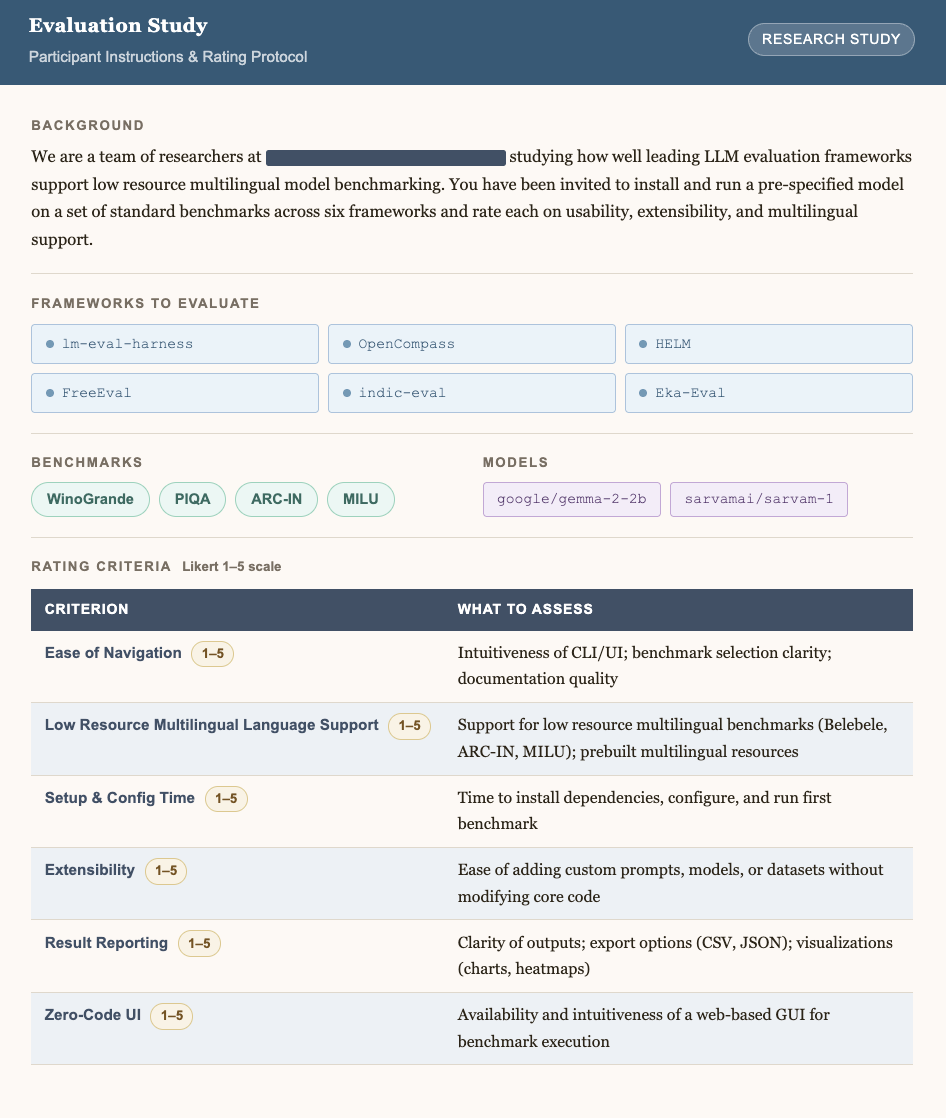}
    \caption{Participants assessment form.}
    \label{fig:study}
\end{figure*}

\begin{figure*}[b]
    \centering
    \includegraphics[width=1\linewidth]{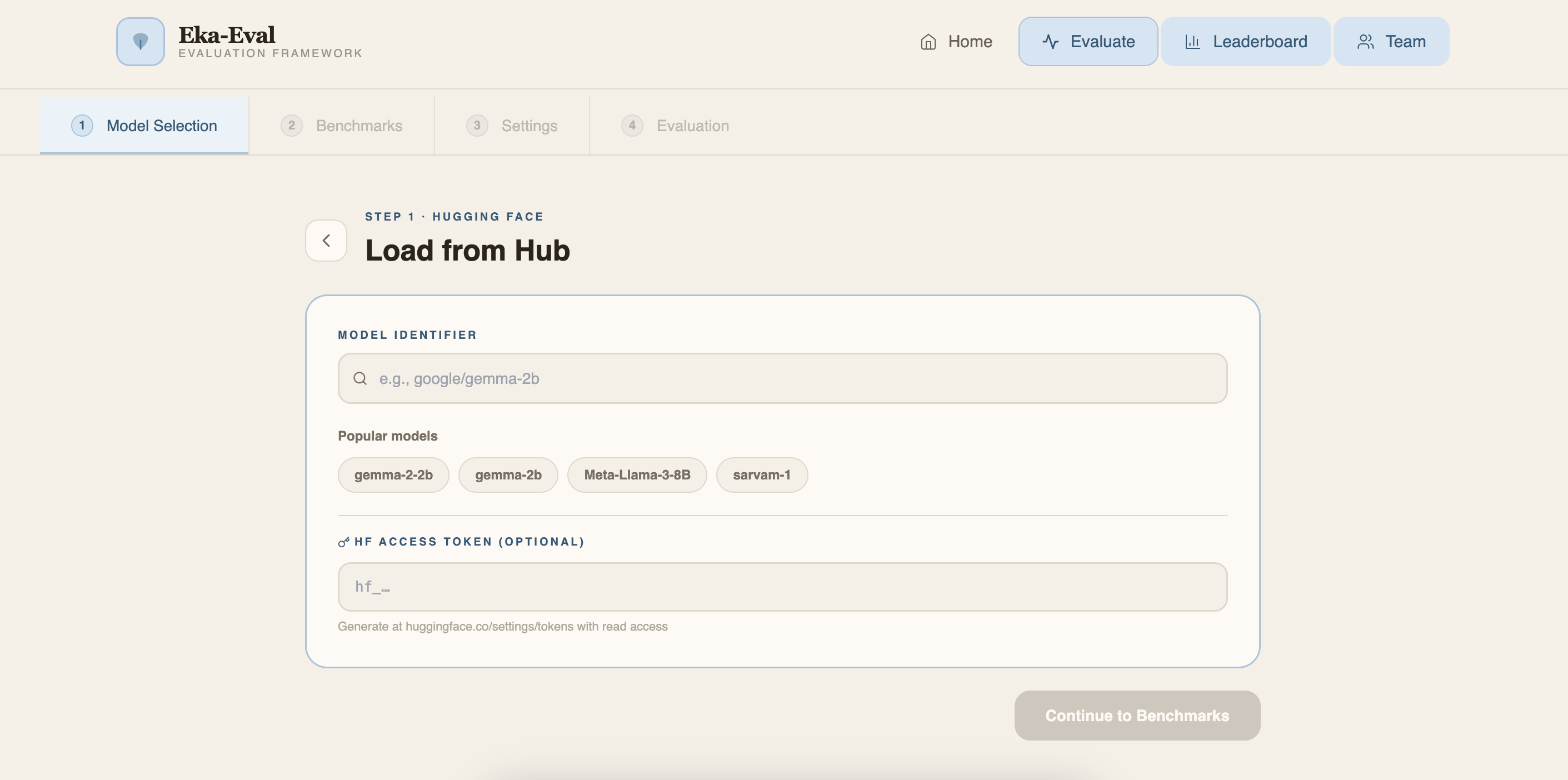}
    \caption{ModelHub selection panel to choose any HuggingFace/Local models for benchmark evaluation.}
    \label{fig:model-hub}
\end{figure*}

\begin{figure*}[b]
    \centering
    \includegraphics[width=1\linewidth]{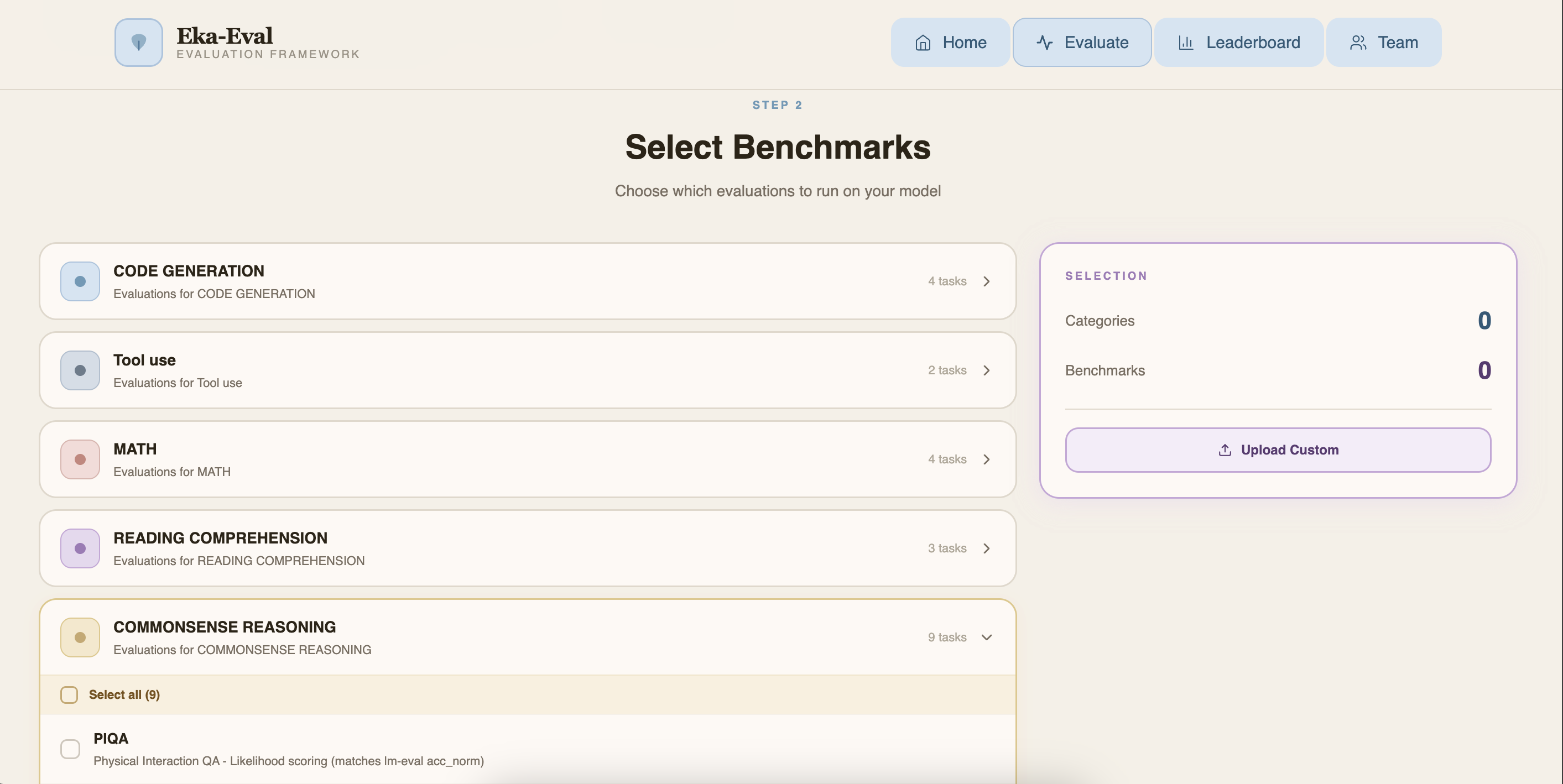}
    \caption{Benchmark selection dashboard categorized by domain (e.g., Math, Code Generation, and other benchmarks.}
    \label{fig:benchmarks2}
\end{figure*}

\begin{figure*}[h]
    \centering
    \includegraphics[width=1\linewidth]{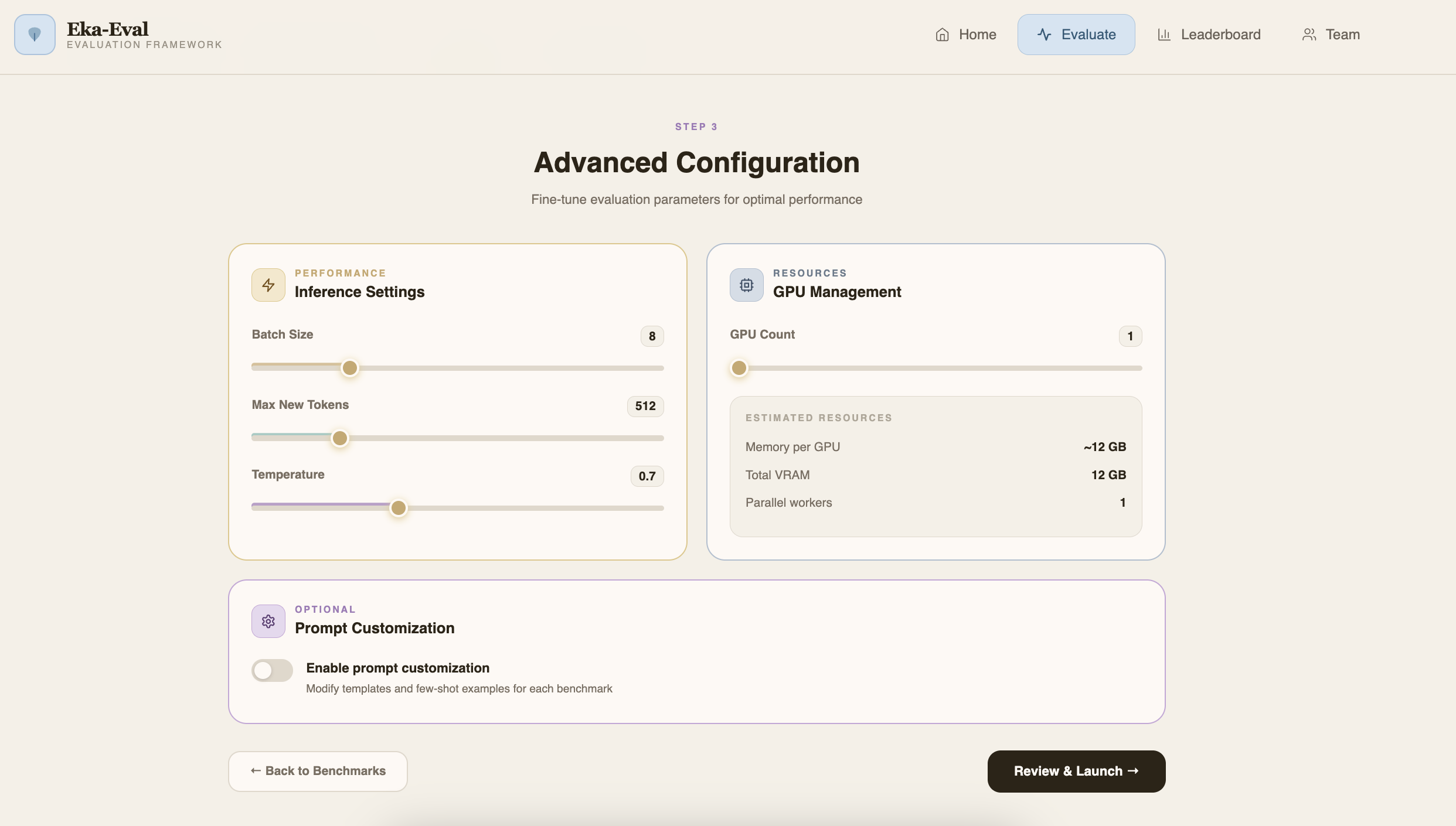}
    \caption{Advanced configuration panel for adjusting inference parameters (temperature, batch size) and managing GPU resources.}
    \label{fig:adv}
\end{figure*}

\begin{figure*}[h]
    \centering
    \includegraphics[width=1\linewidth]{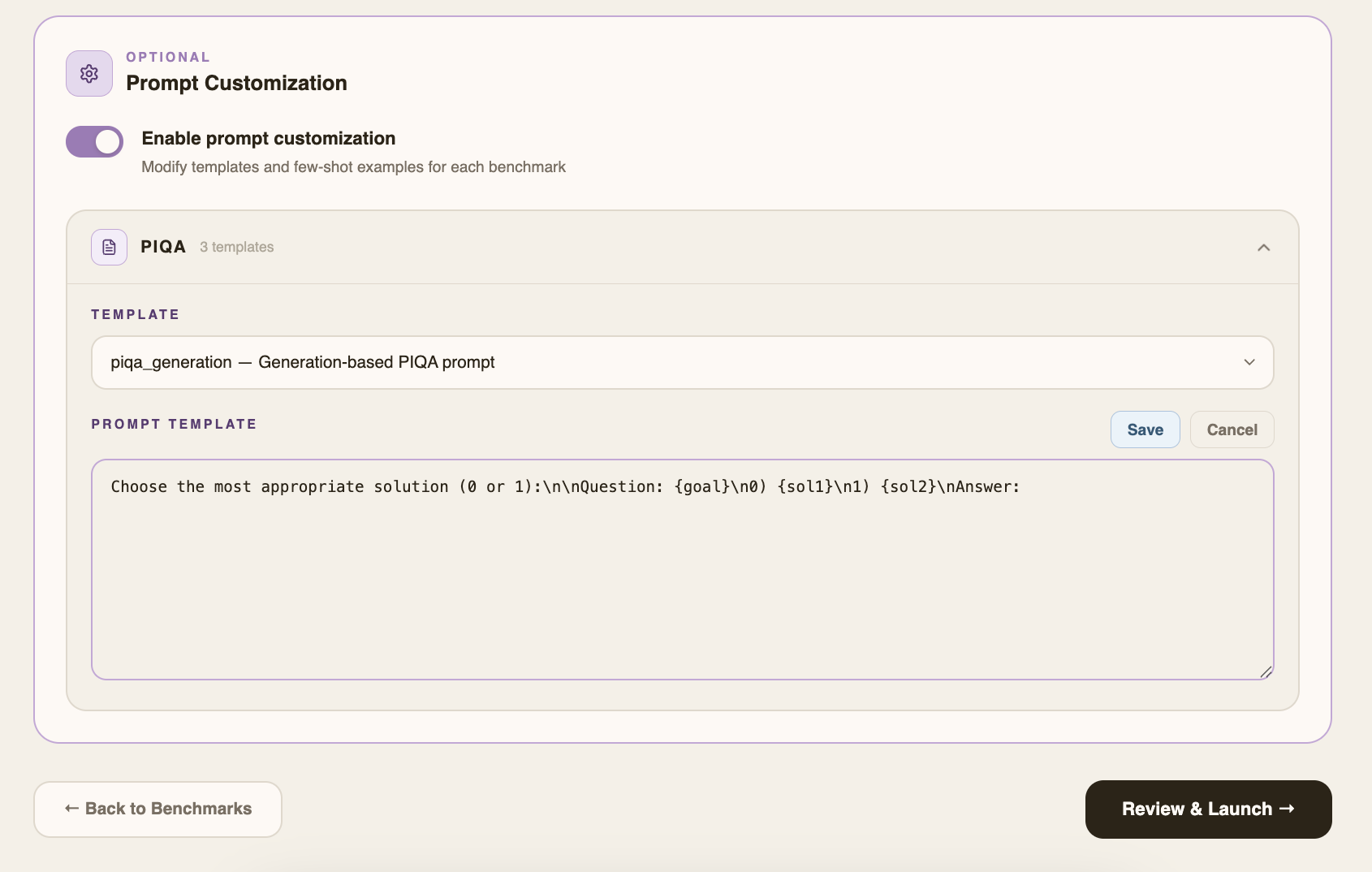}
    \caption{The prompt customization interface, allowing users to modify templates and injection points for specific tasks.}
    \label{fig:promptemp}
\end{figure*}

\begin{figure*}[h]
    \centering
    \includegraphics[width=1\linewidth]{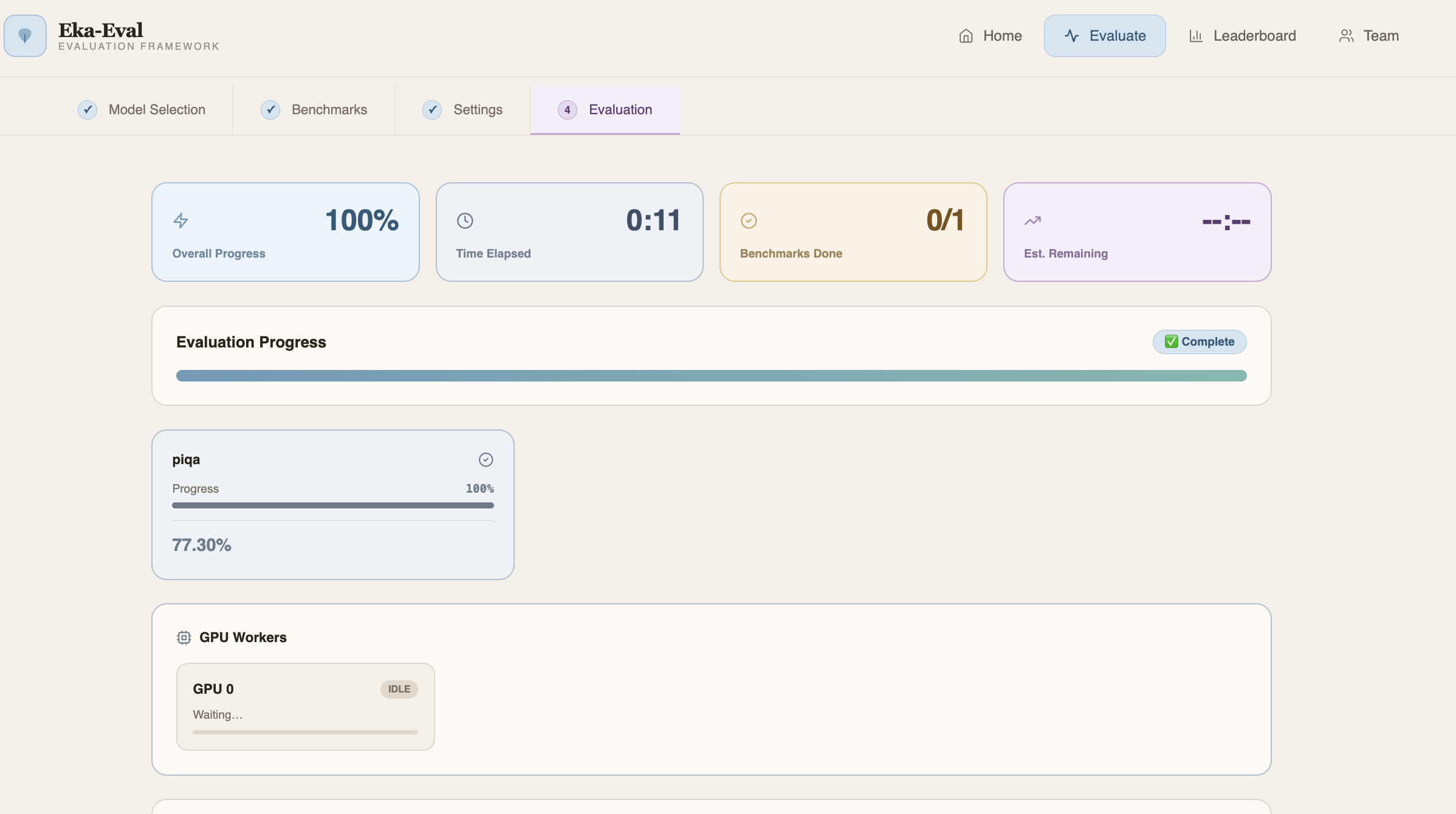}
    \caption{Post-evaluation summary showing overall progress, specific benchmark scores, and options to export results or view detailed analysis.}
    \label{fig:live}
\end{figure*}

\begin{figure*}[h]
    \centering
    \includegraphics[width=1\linewidth]{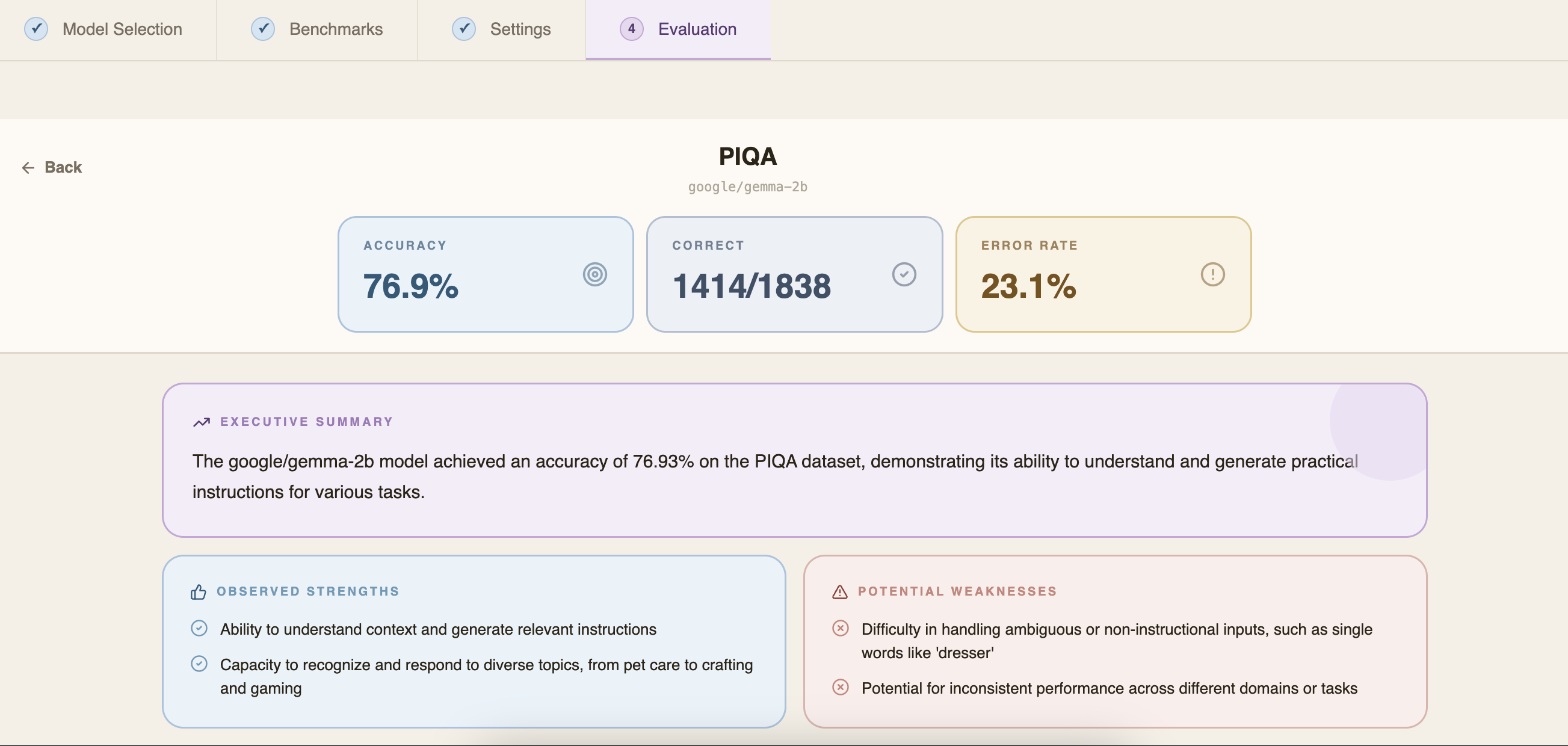}
    \caption{AI diagnosis dashboard providing automated textual insights into model strengths and weaknesses based on performance.}
        \label{fig:summary}
\end{figure*}

\begin{figure*}[h]
        \centering
        \includegraphics[width=1\linewidth]{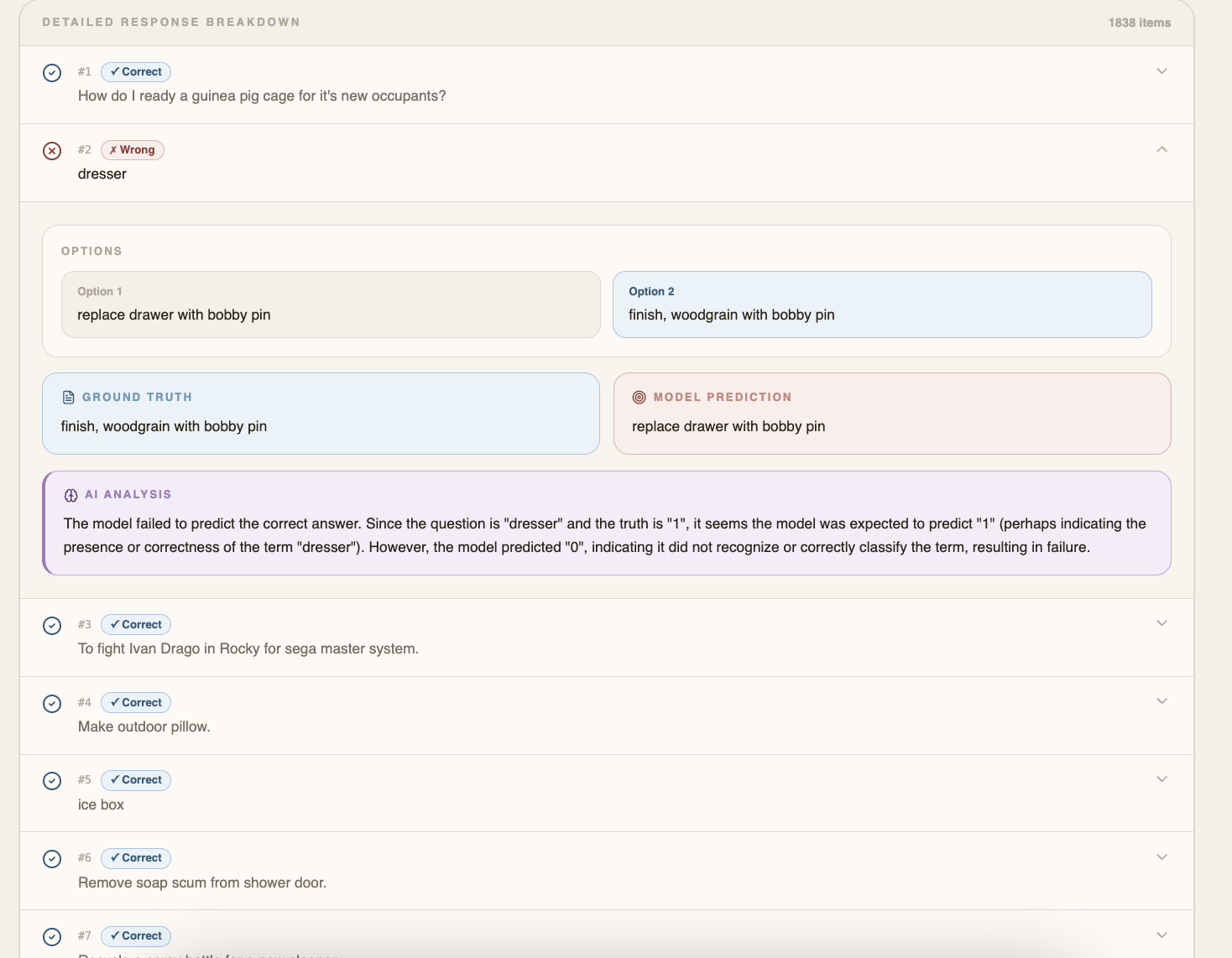}
        \caption{AI diagnosis showing the wrong result of selected benchmark}
        \label{fig:placeholder2}
\end{figure*}

\begin{figure*}[h]
    \centering
    \includegraphics[width=1\linewidth]{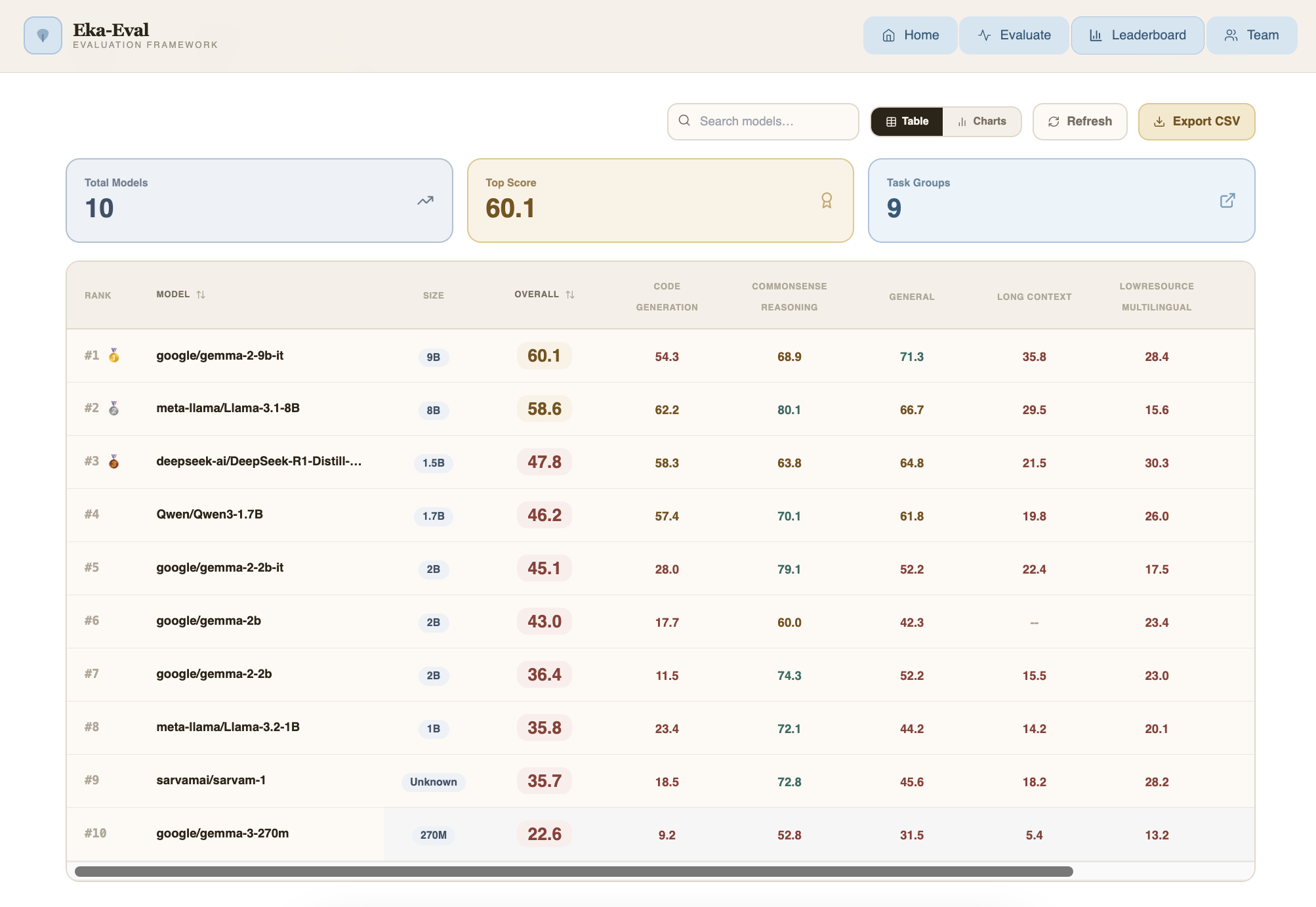}
    \caption{Model Leaderboard through aggregated average scores across multiple benchmarks and through live evaluation}
    \label{fig:leaderboardd}
\end{figure*}

\begin{figure*}[h]
    \centering
    \includegraphics[width=1\linewidth]{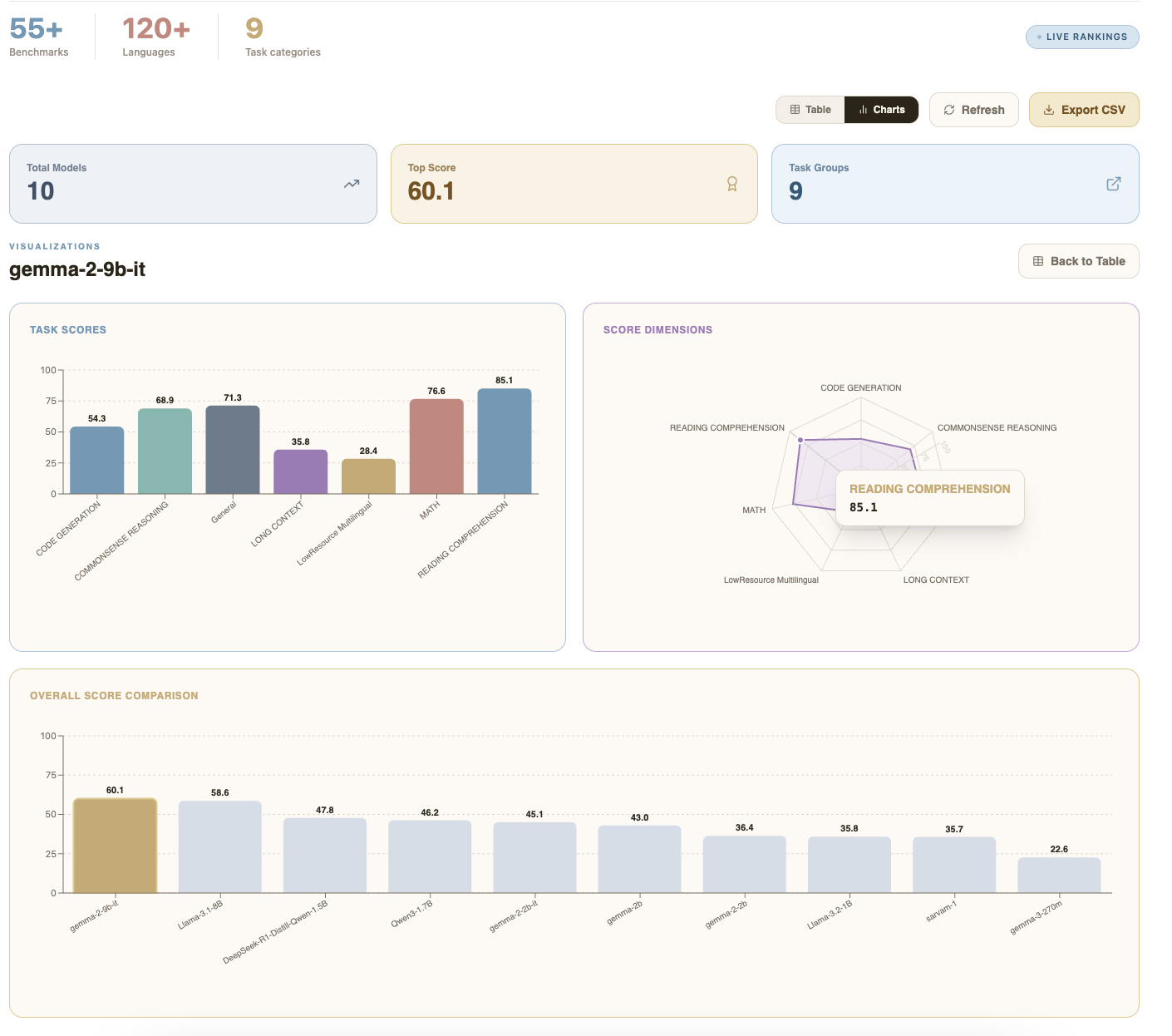}
    \caption{Model Leaderboard featuring multi-dimensional visualizations (radar and bar charts) for comparative analysis.}
    \label{fig:visuals23}
\end{figure*}

\begin{figure*}[h]
    \centering
    \includegraphics[width=1\linewidth]{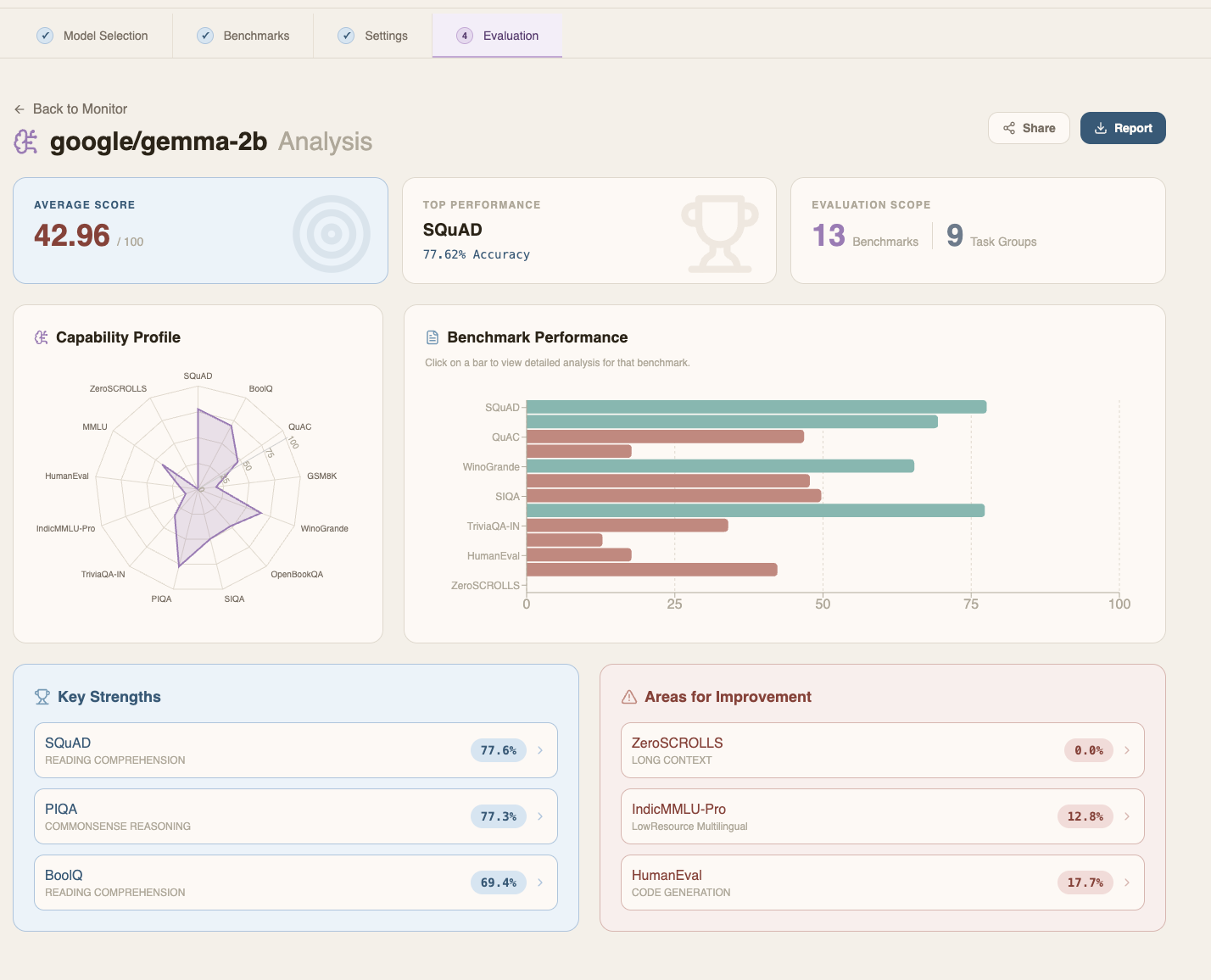}
    \caption{Model Key strengths and weaknesses featuring multi-dimensional visualizations along with benchmarks performance analysis for improvement.}
    \label{fig:visuals32}
\end{figure*}

% \begin{figure*}[h]
%     \centering
%     \includegraphics[width=1\linewidth]{ui2.png}
%     \caption{Benchmark selection dashboard categorized by domain (e.g., Math, Code Generation, and Other Benchmarks).}
%     \label{fig:ui-benchmarks}
% \end{figure*}

\paragraph{The Benchmark Selection Dashboard} (Figure~\ref{fig:benchmarks2}) allows users to build their own evaluation suite by simply toggling and selecting the benchmarks available and also adding custom benchmarks.

% \begin{figure*}[h]
%     \centering
%     \includegraphics[width=1\linewidth]{ui4.png}
%     \caption{Advanced configuration panel for adjusting inference parameters (temperature, batch size) and managing GPU resources.}
%     \label{fig:ui-gpu}
% \end{figure*}

\noindent For advanced control over the evaluation process, 
\textbf{Configuration Panel} (as shown in Figure~\ref{fig:adv}) offers adjustable settings such as temperature, top-p and batch size. It also includes a built-in GPU resource manager, allowing users to allocate specific or multi-GPU support for evaluation.

% \begin{figure*}[h]
%     \centering
%     \includegraphics[width=1\linewidth]{ui5.png}
%     \caption{The prompt customization interface, allowing users to modify templates and injection points for specific tasks.}
%     \label{fig:ui-prompts}
% \end{figure*}

\noindent A key advantage of the web user interface is the \textbf{Prompt Customization Engine} (Figure~\ref{fig:promptemp}). It lets users visually edit system prompts, few-shot templates, and variable fields, GPU allocation, thus removing the need to manually modify JSON configuration files.

% \begin{figure*}[h]
%     \centering
%     \includegraphics[width=1\linewidth]{ui6.png}
%     \caption{Post-evaluation summary showing overall progress, specific benchmark scores, and options to export results or view detailed analysis.}
%     \label{fig:ui-eval}
% \end{figure*}

\paragraph{The Evaluation Dashboard} Figure~\ref{fig:live} shows real time progress during inference including a live console and, once finished, it provides a summary of the model’s performance with options to download logs or explore detailed analysis.

% \begin{figure*}[h]
%     \centering
%     \includegraphics[width=1\linewidth]{ui7.png}
%     \caption{AI diagnosis dashboard providing automated textual insights into model strengths and weaknesses based on performance.}
%     \label{fig:ui-ai}
% \end{figure*}

\paragraph{AI Diagnosis Dashboard} uses \texttt{LLama 3.3-70b} model to analyze the evaluation logs, generating summaries of the model's observed strengths (e.g., `Strong logical coherence') and weaknesses (e.g., `Hallucination in low-resource languages') as example shown in Figure~\ref{fig:summary}.

% \begin{figure*}[h]
%     \centering
%     \includegraphics[width=1\linewidth]{ui8.png}
%     \caption{AI diagnosis of the wrong result of benchmark selected}
%     \label{fig:ui-ai2}
% \end{figure*}

\noindent Users can also inspect \textbf{specific failure cases} (Figure~\ref{fig:placeholder2}). Selecting a failed benchmark item shows the prompt, the model’s wrong answer, and an AI-generated explanation of why the error likely occurred, making it easy to spot and fix issues.

% \begin{figure*}[h]
%     \centering
%     \includegraphics[width=1\linewidth]{ui11.png}
%     \caption{Model Leaderboard featuring multi-dimensional visualizations (radar and bar charts) for comparative analysis.}
%     \label{fig:ui-leaderboard}
% \end{figure*}

\noindent Finally, the \textbf{Model Leaderboard} Figure~\ref{fig:leaderboardd}, Figure~\ref{fig:visuals23} and Figure~\ref{fig:visuals32} provides a centralized, dynamic interface for comparative analysis. Unlike static evaluation tables, it aggregates results from all local runs securely. Researchers can filter outputs by model size, language, or specific benchmark, and generate side-by-side multi-dimensional radar charts. To maintain integrity, the leaderboard includes version tracking for datasets and prevents score manipulation by linking directly to the immutable evaluation logs generated by the backend engine. 

%%%
% \begin{figure*}[h]
%     \centering
%     \includegraphics[width=1\linewidth]{leaderboard.png}
%     \caption{Model Leaderboard through dynamic evaluation}
%     \label{fig:leaderboard}
% \end{figure*}

\section{Interactive CLI}
\label{sec:appendix-cli}
\subsection{CLI Demonstration}

The interactive CLI of the \textsc{Eka-Eval} framework is shown below, which guides users through model selection and evaluation setup. Simplifying benchmarking workflows, it is accessible to both researchers and developers.

% \begin{tcolorbox}[title=\textsc{Eka-Eval} CLI: Available Benchmark Task Groups]
% \begin{verbatim}
% --- Available Benchmark Task Groups ---
% 1.  Code Generation
% 2.  Tool Use
% 3.  Math
% 4.  Reading Comprehension
% 5.  Commonsense Reasoning
% 6.  World Knowledge
% 7.  Long Context
% 8.  General
% 9.  Low-Resource Multilingual
% 10. All Task Groups

% Select task group #(s) 
% (e.g., '1', '1 3', 'ALL'):
% \end{verbatim}
% \end{tcolorbox}

\begin{figure*}[h]
    \centering
    \includegraphics[width=1\textwidth, height=6cm, keepaspectratio]{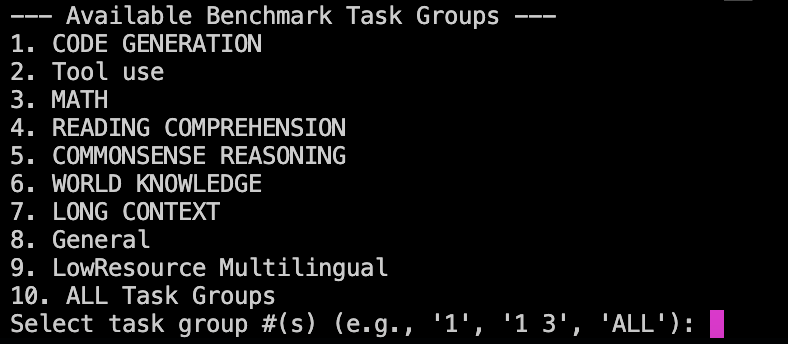}
    \caption{Available benchmark groups of the \textsc{Eka-Eval} framework.}
    \label{fig:tasks}
\end{figure*}

\noindent As per Figure~\ref{fig:tasks}, users are prompted to select high-level task groups (e.g., Reading Comprehension) during CLI setup. This enables fine-grained benchmarking organization and streamlined selection.
\begin{figure*}[h]
    \centering
    \includegraphics[width=1\linewidth,height=6cm, keepaspectratio]{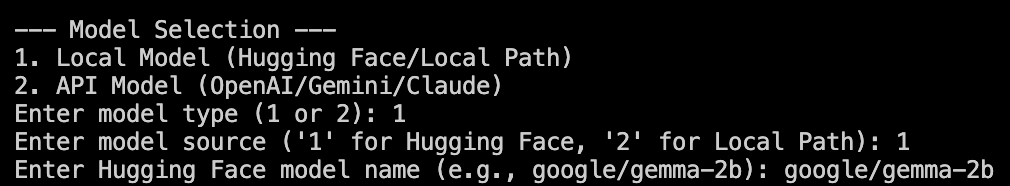}
    \caption{Model selection in the \textsc{Eka-Eval} framework.}
    \label{fig:placeholder}
\end{figure*}

\begin{figure*}[h]
    \centering
    \includegraphics[width=1\linewidth]{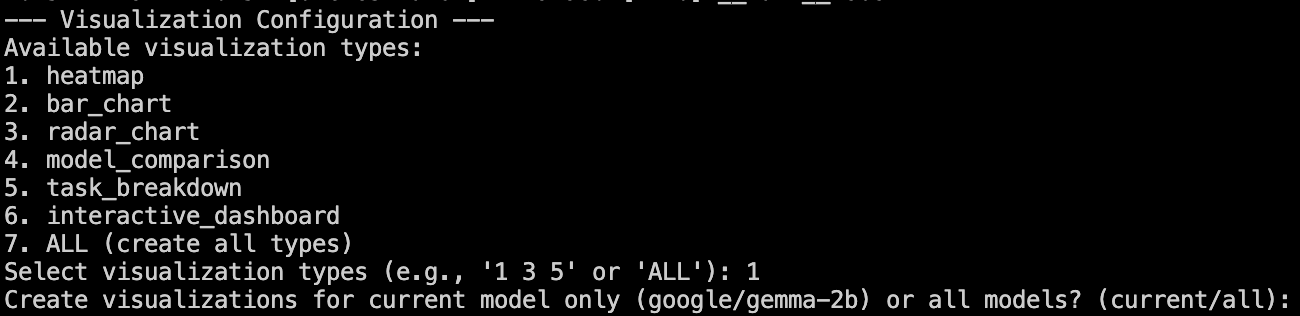}
    \caption{Interactive and user-friendly visualisation setup in \textsc{Eka-Eval}.}
    \label{fig:placeholder1}
\end{figure*}

\begin{figure*}[h]
    \centering
    \includegraphics[width=1\textwidth]{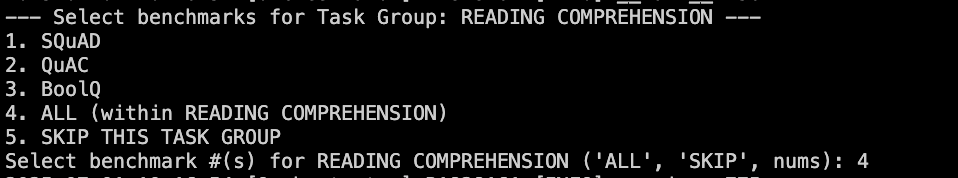}
    \caption{Subtask selection within a task group.}
    \label{fig:sub}
\end{figure*}

\begin{figure*}[h]
    \centering
    \includegraphics[width=0.9\textwidth]{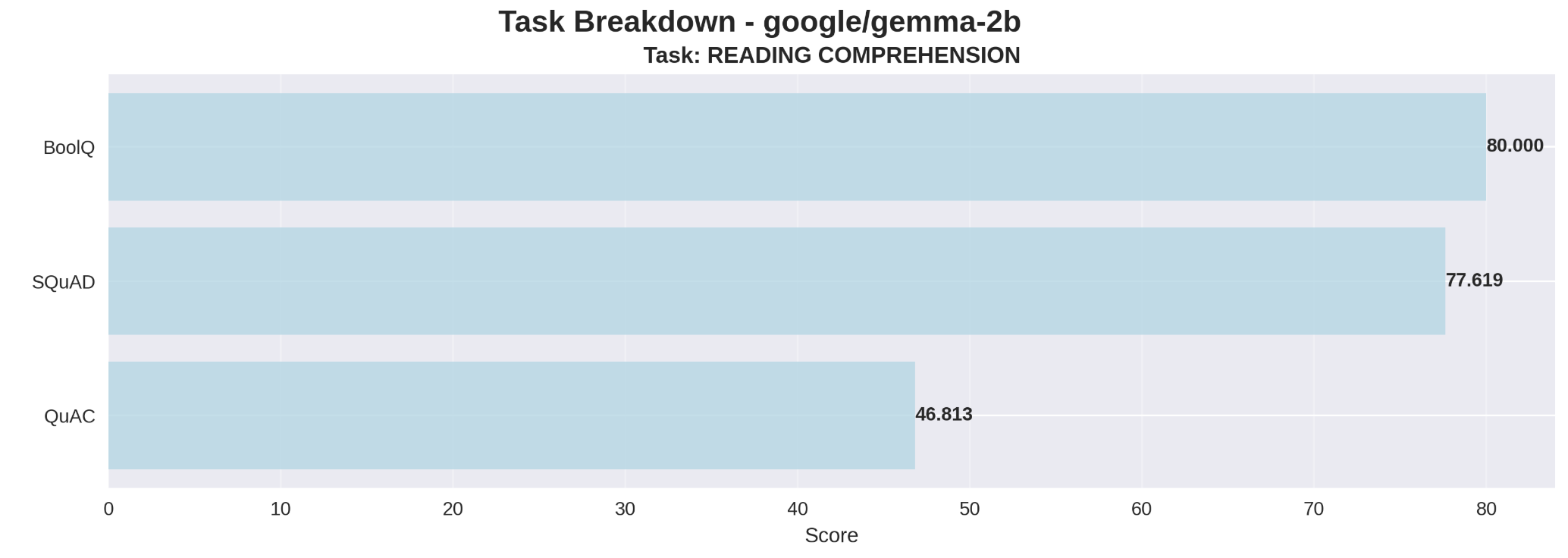}
    \caption{Bar chart visualisation on Reading Comprehension benchmarks}
    \label{fig:bar-chart}
\end{figure*}

% \noindent \textsc{Eka-Eval} supports local HuggingFace models and API-based models like Sarvam, Gemma, OpenAI, Claude, and Gemini. Users interactively select model source and configuration through CLI (see Figure~\ref{fig:model-selection}).

\noindent After selecting a task group, users choose specific benchmarks such as SQuAD, BoolQ, or QuAC for focused evaluation within that domain (see Figure~\ref{fig:sub}).

\begin{figure*}[h]
    \centering
    \includegraphics[width=1\linewidth]{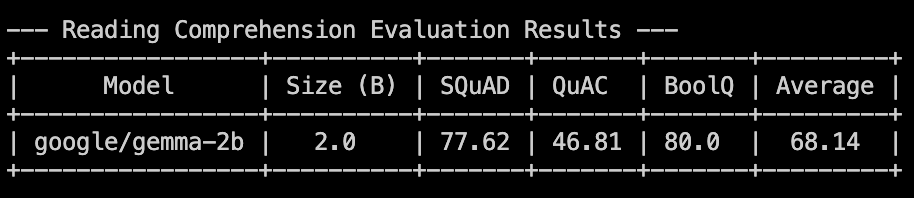}
    \caption{Consolidated evaluation results table.}
    \label{fig:results-table}
\end{figure*}

\noindent The CLI displays final benchmark scores for each model in tabular format, including per-task and average scores. Results are also exported as CSV (see Figure~\ref{fig:results-table}).

% \noindent As per Figure~\ref{fig:viz-setup}) framework allows users to generate multiple types of visualizations: bar charts, heatmaps, radar plots that are based on completed evaluations. 

\noindent Figure~\ref{fig:bar-chart} shows performance breakdown across sub-tasks like BoolQ, SQuAD, and QuAC. It provides intuitive insight into strengths and weaknesses of the model.

\section{User Study Protocol}
\label{sec:guidelines}

Participants received a standardized evaluation protocol 
and rating form (Figures~\ref{fig:rating} and~\ref{fig:study}) 
to ensure consistency across all framework evaluations. 
Each participant was permitted to consult official 
documentation but was prohibited from seeking external 
technical assistance beyond publicly available framework 
resources. Explicit guidelines specified permitted and 
prohibited actions to minimize evaluation bias.

\section{Extensibility and Customization}
\textsc{Eka-Eval} emphasizes extensibility through a low-code plugin architecture that enables easy integration of new benchmarks and custom metrics with minimal modifications, managed via a hierarchical JSON configuration system that supports complex setups such as parameter sweeps and prompt variations. The framework provides a zero-code UI that enables real-time, interactive adjustment of inference parameters (e.g., temperature, batch size) and evaluation strategies, eliminating the need for users to edit the configuration files directly.

\section{Computation Requirement and Budget}
For computational infrastructure, experiments were carried out on four NVIDIA Tesla V100 32 GB GPUs, with an estimated cost of \$5,431.20 per month based on Google Cloud Platform (GCP) \footnote{\url{https://cloud.google.com/products/calculator}} Calculator pricing.

%// with an estimated cost of $7,192.00 per month based on Google Cloud
%Platform (GCP) 20 Calculator pricing.//

\end{document}